\documentclass[journal]{IEEEtran}
\usepackage{times}
\usepackage{epsfig}
\usepackage{graphicx}
\usepackage{amsmath}
\usepackage{amssymb}
\usepackage{multirow}
\usepackage{subcaption}
\usepackage{soul}
\usepackage{color}
\usepackage{url}
\usepackage[switch]{lineno}
\usepackage[compact]{titlesec}
\usepackage[backend=bibtex,bibstyle=ieee,citestyle=numeric-comp]{biblatex}
\bibliography{refs}
\begin{document}
%
\title{Divergence Regulated Encoder Network for Joint Dimensionality Reduction and Classification}
%
%
%

\author{Joshua~Peeples,~\IEEEmembership{Student Member,~IEEE,}
Sarah Walker,~\IEEEmembership{Student Member,~IEEE,}
Connor McCurley,~\IEEEmembership{Student Member,~IEEE,}
Alina Zare,~\IEEEmembership{Senior Member,~IEEE,}
James Keller,~\IEEEmembership{Life~Fellow,~IEEE, }
and~Weihuang~Xu,~\IEEEmembership{Student Member,~IEEE}
\thanks{J. Peeples, S. Walker, C. McCurley, A. Zare, and W. Xu are with the Department
of Electrical and Computer Engineering, University of Florida, Gainesville, FL, 32608 USA. }
\thanks{J. Keller is with the Department of Electrical Engineering and Computer Science University of Missouri, Columbia, MO, 65211, USA.}
\thanks{Corresponding author: J. Peeples, email: jpeeples@ufl.edu}%
\thanks{Manuscript submitted September 12, 2021. Major revision submitted December 23, 2021. Minor revision submitted February 01, 2022. Accepted March 1, 2022.}}

\markboth{ACCEPTED TO IEEE GEOSCIENCE AND REMOTE SENSING LETTERS}%
{Shell \MakeLowercase{\textit{et al.}}: Bare Demo of IEEEtran.cls for Journals}


\maketitle

\begin{abstract}
  Feature representation is an important aspect of remote-sensing based image classification.  While deep convolutional neural networks are able to effectively amalgamate information, large numbers of parameters often make learned features inscrutable and difficult to transfer to alternative models.  In order to better represent statistical texture information for remote-sensing image classification, in this paper, we investigate performing joint dimensionality reduction and classification using a novel histogram neural network. Motivated by a popular dimensionality reduction approach, t-Distributed Stochastic Neighbor Embedding (t-SNE), our proposed method incorporates a classification loss computed on samples in a low-dimensional embedding space.  We compare the learned sample embeddings against coordinates found by t-SNE in terms of classification accuracy and qualitative assessment. We also explore use of various divergence measures in the t-SNE objective. The proposed method has several advantages such as readily embedding out-of-sample points and reducing feature dimensionality while retaining class discriminability.  Our results show that the proposed approach maintains and/or improves classification performance and reveals characteristics of features produced by neural networks that may be helpful for other applications. 
\end{abstract}


\IEEEpeerreviewmaketitle

\section{Introduction}
\label{sec:intro}
\IEEEPARstart{R}{ecently}, deep convolutional neural network (DCNN)-based approaches have shown remarkable performance on remote sensing tasks \cite{shi2022remote}.  State-of-the-art image classification networks often contain thousands or even millions of parameters which need to be learned during training.  While it has been argued in the literature that high-dimensionality (data, parameterization, model complexity) is more amenable for classification \cite{Gorban2019BlessingofDimensionality,Huang2019UnderstandingGeneralization}, methods taking advantage of high-dimensionality are often inscrutable.  One approach for handling the nuances associated with high-dimensionality is simply to reduce the dimensionality of the data/model.   

\textit{Dimensionality Reduction} (DR) has proven to be a critical tool in many remote sensing applications, such as: hyperspectral land-cover classification \cite{Duan2021DimRedHSI,Luo2021DimRedHSI}, multi-modality sensor fusion \cite{Hong2018CommonSubspaceLearningHSI}, co-registration and alignment \cite{Hong2019LearnableManifoldAlignment}, redundancy removal, memory usage minimization, and reduction of the effects of the Curse of Dimensionality \cite{VanDerMaaten2009DRReview}.  The goal of DR can be posed as discovering intrinsic, lower-dimensional features from the data which meet an overarching objective, such as: preserving variance, finding compressed representations of data, maintaining global or local structure or promoting class discriminability in the embedding space \cite{VanDerMaaten2009DRReview,Bengio2014RepLearningReview, Thorstensen2009ManifoldThesis}.  Most studies perform classification or regression after applying unsupervised dimensionality reduction.  However, it has been shown that there are advantages to learning the low-dimensional representations and classification/regression models simultaneously \cite{Chao2019RecentAdvancesSupervisedDimRed,Rish2008SupDimRedGLM}.  Specific to classification, the goal of DR is to discover embedding functions that take data from the input feature space and transform it into a lower-dimensional coordinate system or \textit{latent space}. Ideally, the low-dimensional features capture ``useful" properties of the data while enforcing constraints such as topological ordering and class separability \cite{Vural2018StudySupervisedManifoldLearning}.
\begin{figure*}[htb]
  \centering
  \includegraphics[width=.81\textwidth]{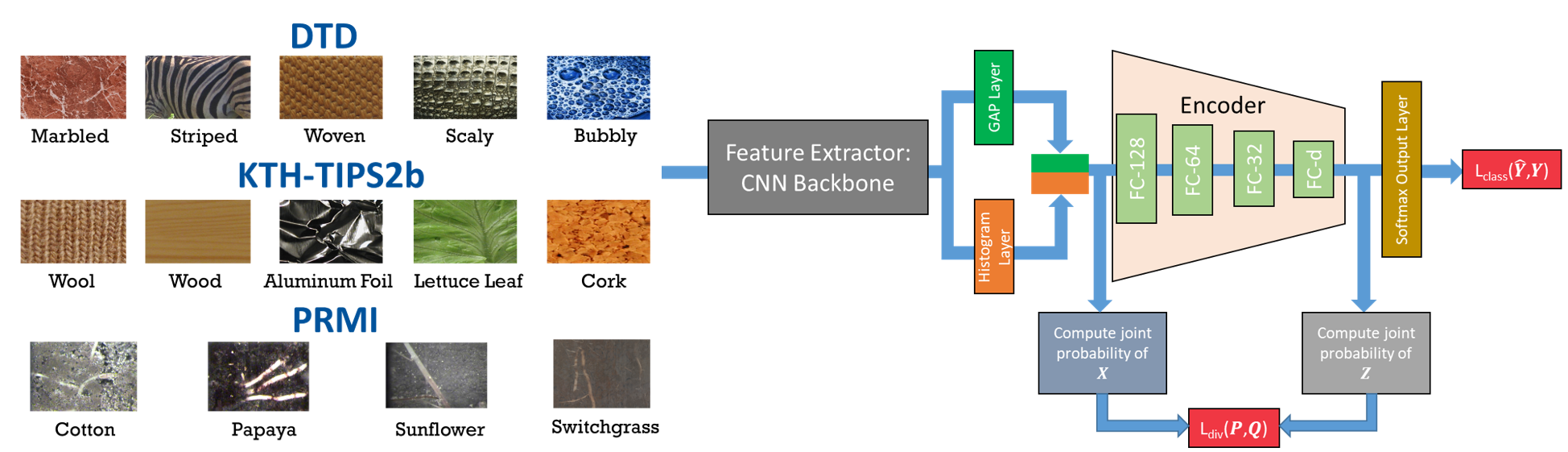}
  \caption{Architecture for the proposed divergence regulated encoder networks (DRENs).}
  \label{fig:model}
\end{figure*}

The representation of data is a critical factor in determining machine learning classification performance \cite{Bengio2014RepLearningReview}.  For images, one can generally represent information through color, shape, and texture features \cite{hiremath2007content}. As shown in the literature \cite{liu2019bow}, texture features often serve as the most powerful descriptor of the three. As currently constructed, DCNNs cannot directly model statistical texture information without the use of extra layers and parameters. To capture statistical texture features in DCNNs, \textit{histogram layer} networks \cite{peeples2020histogram} were introduced to characterize the distribution of features in the model. We hypothesize that incorporating statistical texture features into our proposed approach will assist in providing meaningful representations of the data for classification. 

Based on these motivations, we introduce a neural classification approach which inherently learns compressed feature representations optimized for class discriminability.  
The contributions of this work are summarized as the following:
\begin{itemize}
  \item  We propose a neural classification scheme which finds discriminative, low-dimensional representations of the input data.  The network learns from a t-SNE based objective, yet outperforms the data representations found by classical t-SNE in terms of classification accuracy.
  \item The proposed method is constructed from a histogram layer network backbone which extracts highly descriptive texture features. 
  \item The proposed approach can readily embed out-of-sample points, allowing it to be applied universally to other image classification datasets. 
\end{itemize}


\section{Methods}
\label{sec:methods}
\subsection{Proposed Approach}
In order to jointly perform classification and DR, our objective function is comprised of two terms: classification loss, $L_{class}$ , and a divergence measure, $L_{div}$:
\begin{equation}
    L_{total} = (1-\lambda)L_{class}(\hat{\mathbf{Y}},\mathbf{Y}) +\lambda L_{div}(\mathbf{P},\mathbf{Q})
\end{equation}
where $\mathbf{P} \in \mathbb{R}^{N \times N}$ is the joint probability distribution on the high dimensional input features $\mathbf{X} \in \mathbb{R}^{N \times D}$ for $N$ samples in a mini-batch, $\mathbf{Q} \in \mathbb{R}^{N \times N}$ is the joint probability distribution on the lower dimensional embeddings $\mathbf{Z} \in \mathbb{R}^{N \times d}$, $\hat{\mathbf{Y}} \in \mathbb{R}^{N \times C}$ are the sample predictions for $C$ classes, and $\mathbf{Y} \in \mathbb{R}^{N \times C}$ are the true class labels. We add a weight, $\lambda$, to leverage the contribution of each term to the overall objective. We define $L_{class}$ as the popular cross-entropy loss:
\begin{equation}
   L_{class}(\hat{\mathbf{Y}},\mathbf{Y}) = -\frac{1}{N}\sum_{i=1}^{N}\sum_{c=1}^{C}y_{ic}\log\hat{y}_{ic}
\end{equation}

For t-SNE, Kullback-Leibler (KL) divergence is typically used to minimize the divergence between the higher and lower dimensional joint probability distributions:
\begin{equation}
    L_{div}(\mathbf{P},\mathbf{Q}) = \sum_{i=1}^{N}\sum_{j=1}^{N} p_{ij}\log\frac{p_{ij}}{q_{ij}}
\end{equation}
Following the standard t-SNE \cite{VanDerMaaten2008Visualizing} implementation, we used a normalized radial basis kernel and Student t-distribution with a single degree of freedom to estimate the neighbor probability matrices $\mathbf{P}$ and $\mathbf{Q}$ in the high and low dimensional feature spaces, respectively. Our approach is agnostic to the selection of a divergence measure. Hence, we used other divergence measures such as Wassertein-$1$ (\textit{i.e.}, earth mover's distance) and Renyi's divergence with $\alpha = 0.5$. There was not a statistically significant difference in the classification performance of our model. Therefore, results presented in this work are representative of using KL divergence.

\begin{table*}[htb] 
\centering
\caption{Encoder experiments performance of the different variations of DREN for KTH-TIPS-2b. The best average result for each model, $d$, and $\lambda$ is bolded. The performance and dimensionality without an encoder are shown in the headings. For clarity, the standard deviations are not shown.}
\label{tab:KTH_DREN}
\begin{tabular}{|c|ccccc|ccccc|}
\hline
                   & \multicolumn{5}{c|}{Resnet18 ($D = 512$, 78.77 $\pm$ 2.86)}                                                                                                                    & \multicolumn{5}{c|}{HistRes18 ($D = 1024$, 77.96 $\pm$ 2.12)}                                                                                                                 \\ \hline
Weight ($\lambda$) & \multicolumn{1}{c|}{$d = 2$}             & \multicolumn{1}{c|}{$d = 3$}             & \multicolumn{1}{c|}{$d = 4$}             & \multicolumn{1}{c|}{$d = 8$}             & $d = 16$            & \multicolumn{1}{c|}{$d = 2$}             & \multicolumn{1}{c|}{$d = 3$}             & \multicolumn{1}{c|}{$d = 4$}             & \multicolumn{1}{c|}{$d = 8$}             & $d = 16$                      \\ \hline
$\lambda=0.0$      & \multicolumn{1}{c|}{75.11}          & \multicolumn{1}{c|}{77.55}          & \multicolumn{1}{c|}{78.64}          & \multicolumn{1}{c|}{78.11}          & \textbf{80.37} & \multicolumn{1}{c|}{74.94}          & \multicolumn{1}{c|}{75.84}          & \multicolumn{1}{c|}{\textbf{80.11}} & \multicolumn{1}{c|}{78.09}          & 76.87          \\ \hline
$\lambda = 0.1$    & \multicolumn{1}{c|}{73.38}          & \multicolumn{1}{c|}{76.28}          & \multicolumn{1}{c|}{76.35}          & \multicolumn{1}{c|}{78.09}          & 75.53          & \multicolumn{1}{c|}{\textbf{75.74}} & \multicolumn{1}{c|}{77.74}          & \multicolumn{1}{c|}{79.08}          & \multicolumn{1}{c|}{75.08}          & 74.98          \\ \hline
$\lambda = 0.2$    & \multicolumn{1}{c|}{\textbf{76.91}} & \multicolumn{1}{c|}{78.47}          & \multicolumn{1}{c|}{78.39}          & \multicolumn{1}{c|}{76.81}          & {78.72} & \multicolumn{1}{c|}{74.28}          & \multicolumn{1}{c|}{77.06}          & \multicolumn{1}{c|}{78.05}          & \multicolumn{1}{c|}{78.24}          & \textbf{78.96} \\ \hline
$\lambda = 0.3$    & \multicolumn{1}{c|}{76.41}          & \multicolumn{1}{c|}{\textbf{78.24}} & \multicolumn{1}{c|}{\textbf{78.79}} & \multicolumn{1}{c|}{\textbf{79.50}} & 77.19          & \multicolumn{1}{c|}{75.06}          & \multicolumn{1}{c|}{76.22}          & \multicolumn{1}{c|}{78.58}          & \multicolumn{1}{c|}{77.19}          & 77.50          \\ \hline
$\lambda = 0.4$    & \multicolumn{1}{c|}{71.15}          & \multicolumn{1}{c|}{74.62}          & \multicolumn{1}{c|}{75.82}          & \multicolumn{1}{c|}{77.15}          & 73.61          & \multicolumn{1}{c|}{73.78}          & \multicolumn{1}{c|}{78.16}          & \multicolumn{1}{c|}{77.00}          & \multicolumn{1}{c|}{75.93}          & 77.61          \\ \hline
$\lambda = 0.5$    & \multicolumn{1}{c|}{70.98}          & \multicolumn{1}{c|}{76.79}          & \multicolumn{1}{c|}{77.46}          & \multicolumn{1}{c|}{77.27}          & 78.39          & \multicolumn{1}{c|}{73.19}          & \multicolumn{1}{c|}{\textbf{80.07}} & \multicolumn{1}{c|}{77.74}          & \multicolumn{1}{c|}{75.27}          & 78.22          \\ \hline
$\lambda = 0.6$    & \multicolumn{1}{c|}{71.97}          & \multicolumn{1}{c|}{75.53}          & \multicolumn{1}{c|}{75.65}          & \multicolumn{1}{c|}{\textbf{78.43}} & 77.63          & \multicolumn{1}{c|}{74.01}          & \multicolumn{1}{c|}{76.96}          & \multicolumn{1}{c|}{76.49}          & \multicolumn{1}{c|}{\textbf{79.82}} & 78.45          \\ \hline
$\lambda = 0.7$    & \multicolumn{1}{c|}{73.46}          & \multicolumn{1}{c|}{73.40}          & \multicolumn{1}{c|}{76.41}          & \multicolumn{1}{c|}{78.14}          & 78.77          & \multicolumn{1}{c|}{71.28}          & \multicolumn{1}{c|}{73.59}          & \multicolumn{1}{c|}{79.99}          & \multicolumn{1}{c|}{77.21}          & 76.87          \\ \hline
$\lambda = 0.8$    & \multicolumn{1}{c|}{69.82}          & \multicolumn{1}{c|}{73.78}          & \multicolumn{1}{c|}{75.13}          & \multicolumn{1}{c|}{76.62}          & 77.50          & \multicolumn{1}{c|}{73.99}          & \multicolumn{1}{c|}{75.04}          & \multicolumn{1}{c|}{73.61}          & \multicolumn{1}{c|}{71.55}          & 74.18          \\ \hline
$\lambda = 0.9$    & \multicolumn{1}{c|}{70.29}          & \multicolumn{1}{c|}{70.88}          & \multicolumn{1}{c|}{72.29}          & \multicolumn{1}{c|}{76.77}          & 73.88          & \multicolumn{1}{c|}{71.51}          & \multicolumn{1}{c|}{74.24}          & \multicolumn{1}{c|}{73.55}          & \multicolumn{1}{c|}{78.37}          & 75.63          \\ \hline
$\lambda = 1.0$    & \multicolumn{1}{c|}{17.45}          & \multicolumn{1}{c|}{14.23}          & \multicolumn{1}{c|}{9.41}           & \multicolumn{1}{c|}{16.14}          & 17.34          & \multicolumn{1}{c|}{15.89}          & \multicolumn{1}{c|}{13.87}          & \multicolumn{1}{c|}{18.69}          & \multicolumn{1}{c|}{16.16}          & 15.49          \\ \hline
\end{tabular}
\end{table*}

\begin{table*}[htb] 
\centering
\caption{Encoder experiments performance of the different variations of DREN for DTD. The best average result for each model, $d$, and $\lambda$ is bolded. The performance and dimensionality without an encoder are shown in the headings. For clarity, the standard deviations are not shown.}
\label{tab:DTD_DREN}
\begin{tabular}{|c|ccccc|ccccc|}
\hline
                   & \multicolumn{5}{c|}{Resnet50 ($D = 2048$, 62.79 $\pm$ 0.94)}                                                                                                                   & \multicolumn{5}{c|}{HistRes50 ($D = 4096$, 62.78 $\pm$ 1.12)}                                                                                                               \\ \hline
Weight ($\lambda$) & \multicolumn{1}{c|}{$d = 2$}             & \multicolumn{1}{c|}{$d = 3$}             & \multicolumn{1}{c|}{$d = 4$}             & \multicolumn{1}{c|}{$d = 8$}             & $d = 16$            & \multicolumn{1}{c|}{$d = 2$}             & \multicolumn{1}{c|}{$d = 3$}             & \multicolumn{1}{c|}{$d = 4$}             & \multicolumn{1}{c|}{$d = 8$}             & $d = 16$                      \\ \hline 
$\lambda=0.0$      & \multicolumn{1}{c|}{28.25}          & \multicolumn{1}{c|}{\textbf{45.49}} & \multicolumn{1}{c|}{49.74}          & \multicolumn{1}{c|}{55.46}          & 62.64          & \multicolumn{1}{c|}{\textbf{40.07}} & \multicolumn{1}{c|}{53.06}          & \multicolumn{1}{c|}{55.46}          & \multicolumn{1}{c|}{\textbf{59.74}} & 62.78          \\ \hline
$\lambda = 0.1$    & \multicolumn{1}{c|}{30.34}          & \multicolumn{1}{c|}{44.77}          & \multicolumn{1}{c|}{50.57}          & \multicolumn{1}{c|}{54.96}          & 62.38          & \multicolumn{1}{c|}{39.72}          & \multicolumn{1}{c|}{53.49}          & \multicolumn{1}{c|}{56.21}          & \multicolumn{1}{c|}{59.10}          & 62.65          \\ \hline
$\lambda = 0.2$    & \multicolumn{1}{c|}{\textbf{30.55}} & \multicolumn{1}{c|}{44.23}          & \multicolumn{1}{c|}{\textbf{51.27}} & \multicolumn{1}{c|}{55.50}          & 62.49          & \multicolumn{1}{c|}{38.77}          & \multicolumn{1}{c|}{\textbf{53.08}} & \multicolumn{1}{c|}{56.41}          & \multicolumn{1}{c|}{59.11}          & 62.12          \\ \hline
$\lambda = 0.3$    & \multicolumn{1}{c|}{30.39}          & \multicolumn{1}{c|}{45.10}          & \multicolumn{1}{c|}{50.47}          & \multicolumn{1}{c|}{55.70}          & 63.01          & \multicolumn{1}{c|}{36.54}          & \multicolumn{1}{c|}{52.97}          & \multicolumn{1}{c|}{\textbf{56.60}} & \multicolumn{1}{c|}{59.01}          & 61.61          \\ \hline
$\lambda = 0.4$    & \multicolumn{1}{c|}{29.03}          & \multicolumn{1}{c|}{43.44}          & \multicolumn{1}{c|}{50.77}          & \multicolumn{1}{c|}{56.16}          & 62.58          & \multicolumn{1}{c|}{35.42}          & \multicolumn{1}{c|}{50.61}          & \multicolumn{1}{c|}{56.05}          & \multicolumn{1}{c|}{59.05}          & 61.93          \\ \hline
$\lambda = 0.5$    & \multicolumn{1}{c|}{29.41}          & \multicolumn{1}{c|}{42.80}          & \multicolumn{1}{c|}{50.34}          & \multicolumn{1}{c|}{56.16}          & \textbf{63.16} & \multicolumn{1}{c|}{35.56}          & \multicolumn{1}{c|}{48.54}          & \multicolumn{1}{c|}{56.07}          & \multicolumn{1}{c|}{58.92}          & 62.37          \\ \hline
$\lambda = 0.6$    & \multicolumn{1}{c|}{26.93}          & \multicolumn{1}{c|}{42.83}          & \multicolumn{1}{c|}{50.17}          & \multicolumn{1}{c|}{56.12}          & 62.49          & \multicolumn{1}{c|}{34.51}          & \multicolumn{1}{c|}{48.45}          & \multicolumn{1}{c|}{53.87}          & \multicolumn{1}{c|}{59.30}          & 62.83          \\ \hline
$\lambda = 0.7$    & \multicolumn{1}{c|}{26.24}          & \multicolumn{1}{c|}{40.39}          & \multicolumn{1}{c|}{49.57}          & \multicolumn{1}{c|}{\textbf{56.24}} & 62.57          & \multicolumn{1}{c|}{32.14}          & \multicolumn{1}{c|}{46.30}          & \multicolumn{1}{c|}{52.8}           & \multicolumn{1}{c|}{59.53}          & 62.03          \\ \hline
$\lambda = 0.8$    & \multicolumn{1}{c|}{23.31}          & \multicolumn{1}{c|}{38.89}          & \multicolumn{1}{c|}{44.66}          & \multicolumn{1}{c|}{56.09}          & 62.43          & \multicolumn{1}{c|}{28.19}          & \multicolumn{1}{c|}{43.47}          & \multicolumn{1}{c|}{49.48}          & \multicolumn{1}{c|}{59.54}          & 62.41          \\ \hline
$\lambda = 0.9$    & \multicolumn{1}{c|}{18.04}          & \multicolumn{1}{c|}{33.60}          & \multicolumn{1}{c|}{41.19}          & \multicolumn{1}{c|}{54.32}          & 62.51          & \multicolumn{1}{c|}{23.17}          & \multicolumn{1}{c|}{36.51}          & \multicolumn{1}{c|}{43.16}          & \multicolumn{1}{c|}{57.85}          & \textbf{62.96} \\ \hline
$\lambda = 1.0$    & \multicolumn{1}{c|}{2.97}           & \multicolumn{1}{c|}{3.38}           & \multicolumn{1}{c|}{2.60}           & \multicolumn{1}{c|}{3.43}           & 2.28           & \multicolumn{1}{c|}{2.89}           & \multicolumn{1}{c|}{3.49}           & \multicolumn{1}{c|}{3.29}           & \multicolumn{1}{c|}{2.23}           & 2.92           \\ \hline
\end{tabular}
\end{table*}

\subsection{Implementation}
For our divergence regulated encoder network (DREN), we used two pretrained models as the backbones, ResNet18 and ResNet50 for {Textures under varying Illumination, Pose and Scale (KTH-TIPS-2b) \mbox{\cite{KTHTIPS_Databases}} and Describable Texture Dataset (DTD) \mbox{\cite{cimpoi14describing}}}, respectively.  We added a histogram layer after the last convolutional layer for each of the pretrained models. From these models, we extracted the high dimensional feature vectors, $\mathbf{X}$, for each sample. For the baseline and histogram models used for KTH-TIPS-2b and DTD, $D$ was $512$, $1024$, $2048$, and $4096$ respectively. Our encoder was composed of four fully connected layers of size $128$, $64$, $32$, and $d$. We used ReLU activation functions for the first three layers and passed the $d$-dimensional embedding to a softmax output layer for classification. 

The embeddings, $\mathbf{Z}$, were used to compute $L_{div}$ and to update the weights of the model via backprogation. To create a fair comparison to t-SNE applied to the high dimensional features, we only updated the lower-dimensional embeddings.  Therefore, $\mathbf{P}$ was removed from the computational graph such that only the gradient information from $\mathbf{Q}$ was used. The overall structure of DREN is shown in Figure  \ref{fig:model} and the code for our work is publicly available: \url{https://github.com/GatorSense/DREN}.

\section{Experimental Setup}
Our experiments consisted of two main components: encoder performance and comparisons to t-SNE. For our encoder experiments, we investigated the selection of hyperparameters $\lambda$ and $d$ for the proposed DREN models. We used two texture/material datasets: DTD and KTH-TIPS-2b. DTD \mbox{\cite{cimpoi14describing}} is a collection of textural images annotated with human-centric attributes.  The texture database consists of 5640 images, organized according to 47 categories. The KTH-TIPS-2b \mbox{\cite{KTHTIPS_Databases}} dataset is defined by images of multiple materials under varying pose, scale, and illumination. The database contains images of 11 distinct textures. The typical train/test splits for DTD (train on 80 images per class and test on 40 images per class) and KTH-TIPS-2b (train on three samples, test on one sample) were used for our experiments \cite{liu2019bow}. We held 10\% of the training data for validation and applied each model to the holdout test set. We performed four and ten runs of data splits for KTH-TIPS-2b and DTD respectively. The average classification accuracy is reported across the different folds. We followed a similar training procedure and data augmentation from previous works \cite{peeples2020histogram}, except Adam optimization was used. We also only updated newly added layers (\textit{i.e.}, histogram layers, encoder, output layer) and kept the pretrained weights fixed. The number of bins used in all histogram models was 16.

The weight in the objective term, $\lambda$, was varied from $0$ to $1$ in steps of $0.1$. The embedding dimension, $d$, was set to $2$, $3$, $4$, $8$, and $16$. After our encoder experiments, we compared the embeddings learned by DREN to t-SNE. In order to embed test samples for t-SNE, we used an approach similar to Locally Linear Embedding \cite{Roweis2000LLE}, which assumes that a sample can be represented as a linear combination of its nearest neighbors. Thus, low-dimensional coordinates of unseen test points were found as constrained combinations of their nearest neighbors in the training set. We evaluated each embedding a) qualitatively and b) quantitatively by reporting the classification accuracy of a $K$-NN classifier on the test data ($K=3$). Lastly, we further validated the approach by investigating various CNN backbone architectures. {Additionally, we also show the effectiveness of the approach on a remote sensing dataset, Plant Root Minirhizotron Imagery (PRMI) \mbox{\cite{xu2022prmi}}. We performed species classification on a subset of the data for a total of 4,144 images and four classes. The published training, validation, and test splits were used across five experimental runs.} 


\section{Results and Discussion}
\label{sec:Results}
\begin{figure*}[htb]
    \centering
	\begin{subfigure}{.156\textwidth}{
			\includegraphics[width=\textwidth]{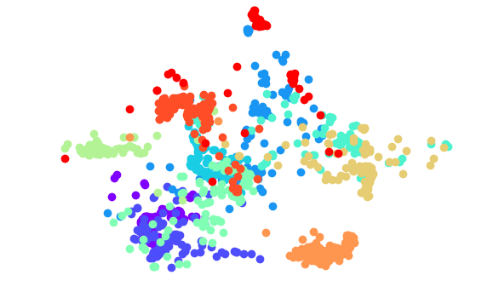}
			\caption{t-SNE}
			\label{fig:KTH_TSNE_embed}
		}
	\end{subfigure}
	\centering
	\begin{subfigure}{.162\textwidth}{
	\includegraphics[width=\textwidth]{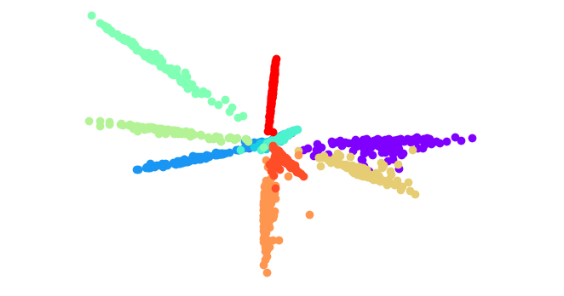}
			\caption{ResNet18}
			\label{fig:KTH_Res_embed}
		}
	\end{subfigure}
	\centering
	\begin{subfigure}{.162\textwidth}{
	\includegraphics[width=\textwidth]{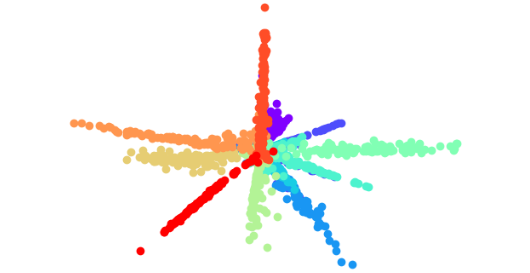}
			\caption{HistRes18}
			\label{fig:KTH_Hist_embed}
		}
	\end{subfigure} 
	\centering
	\begin{subfigure}{.163\textwidth}{
	\includegraphics[width=\textwidth]{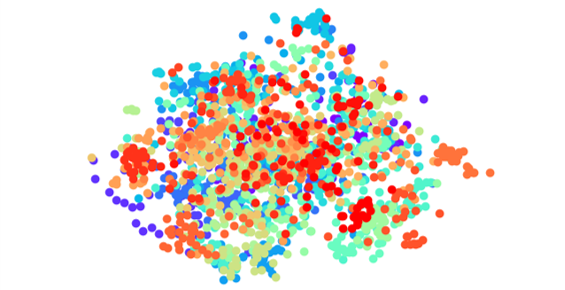}
			\caption{t-SNE}
			\label{fig:DTD_TSNE_embed}
		}
	\end{subfigure}
	\centering
		\begin{subfigure}{.164\textwidth}{
	\includegraphics[width=\textwidth]{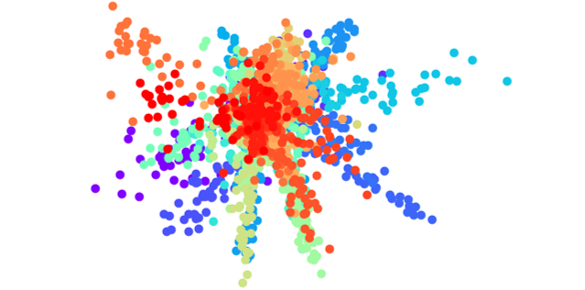}
			\caption{ResNet50}
			\label{fig:DTD_Res_embed}
		}
	\end{subfigure}
	\centering
	\begin{subfigure}{.153\textwidth}{
	\includegraphics[width=\textwidth]{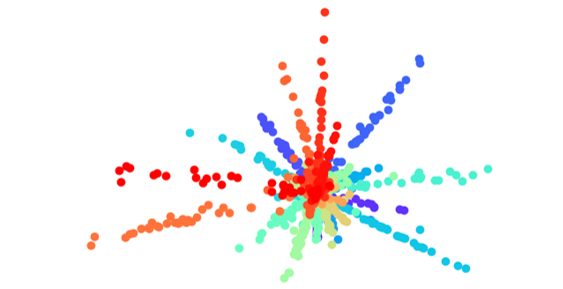}
			\caption{HistRes50}
			\label{fig:DTD_Hist_embed}
		}
	\end{subfigure}
	\caption{2D embeddings of test images from t-SNE out-of-sample approach and different variations of DREN for KTH-TIPS-2b (\ref{fig:KTH_TSNE_embed}-\ref{fig:KTH_Hist_embed}) and DTD (\ref{fig:DTD_TSNE_embed}-\ref{fig:DTD_Hist_embed}). The colors represent different classes from each dataset. In the 2D space, the HistRes features are more separable according to an angular measure, and are also much more compact, \textit{i.e.}, each class mostly varies across a single principal vector, and the principal vectors are well-separated by an angular margin.}
	\label{fig:Embeddings} 
\end{figure*}
\begin{table*}[htb]
\centering
\caption{Test classification performance using $K$-NN ``trained" on embeddings from t-SNE and DREN models. For the DREN models, the weight corresponding to the best average performance is reported. The best average result is bolded.}
\label{tab:KNN}
\begin{tabular}{|c|c|c|c|c|}
\hline
Dataset                      & Embedding Dimension ($d$) & t-SNE              & DREN ResNet               & DREN HistRes             \\ \hline
\multirow{2}{*}{KTH-TIPS-2b} & 2                         & 76.70$\pm$2.88 & \textbf{76.79$\pm$ 3.18} ($\lambda=.2$) & 74.60$\pm$3.47 ($\lambda=.1$) \\ \cline{2-5} 
                             & 3                         & 76.91$\pm$3.40 & 78.18$\pm$3.40 ($\lambda=0$)  & \textbf{79.23$\pm$3.63} ($\lambda=.5$) \\ \hline
\multirow{2}{*}{DTD}         & 2                         & \textbf{43.37$\pm$1.01} & 29.35$\pm$1.22 ($\lambda=.2$)   & 40.44$\pm$1.64 ($\lambda=0$) \\ \cline{2-5} 
                             & 3                         & 43.55$\pm$1.22 & 44.28$\pm$1.57 ($\lambda=0$)   & \textbf{53.02$\pm$1.24} ($\lambda=.1$) \\ \hline
\end{tabular}
\end{table*}
\begin{table*}[htb]
\centering
\caption{{Test classification performance with different backbone architectures for the DREN models. For the DREN models (encoder used), we use equal weighting for each term in the objective ($\lambda = .5$) and the embedding dimension is $d = 16$. The best average result for each model is bolded. For clarity, the standard deviations are not shown. The dimensionalities ($D$) of the base models are shown in the headings.}} 
\label{tab:Architectures}
\begin{tabular}{|c|cllll|c|c|c|c|c|c|}
\hline
Dataset                      & \multicolumn{5}{c|}{Encoder}              & Histogram Layer & \begin{tabular}[c]{@{}c@{}}ResNext50\\ ($D=2048$)\end{tabular} & \begin{tabular}[c]{@{}c@{}}WideResNet50\\ ($D=2048$)\end{tabular} & \begin{tabular}[c]{@{}c@{}}DenseNet121\\ ($D=1024$)\end{tabular} & \begin{tabular}[c]{@{}c@{}}EfficientNet\\ ($D=1280$)\end{tabular} & \begin{tabular}[c]{@{}c@{}}RegNet\\ ($D=400$)\end{tabular} \\ \hline
\multirow{4}{*}{KTH-TIPS-2b} & \multicolumn{5}{c|}{\multirow{2}{*}{No}}  & No              & 57.49                                                          & 74.03                                                             & \textbf{80.49}                                                   & 78.64                                                             & 75.25                                                      \\ \cline{7-12} 
                             & \multicolumn{5}{c|}{}                     & Yes             & 72.29                                                          & 65.80                                                             & \textbf{78.45}                                                   & 74.83                                                             & 74.54                                                      \\ \cline{2-12} 
                             & \multicolumn{5}{c|}{\multirow{2}{*}{Yes}} & No              & 77.02                                                          & 73.84                                                             & \textbf{79.52}                                                   & 76.94                                                             & 76.68                                                      \\ \cline{7-12} 
                             & \multicolumn{5}{c|}{}                     & Yes             & 76.03                                                          & 73.72                                                             & \textbf{81.25}                                                   & 76.66                                                             & 74.56                                                      \\ \hline
\multirow{4}{*}{DTD}         & \multicolumn{5}{c|}{\multirow{2}{*}{No}}  & No              & 26.22                                                          & \textbf{60.18}                                                    & 46.30                                                            & 49.53                                                             & 47.62                                                      \\ \cline{7-12} 
                             & \multicolumn{5}{c|}{}                     & Yes             & 46.72                                                          & \textbf{57.67}                                                    & 55.19                                                            & 45.76                                                             & 54.58                                                      \\ \cline{2-12} 
                             & \multicolumn{5}{c|}{\multirow{2}{*}{Yes}} & No              & 53.69                                                          & 53.32                                                             & \textbf{57.48}                                                   & 52.88                                                             & 53.77                                                      \\ \cline{7-12} 
                             & \multicolumn{5}{c|}{}                     & Yes             & 56.86                                                          & 56.06                                                             & \textbf{60.91}                                                   & 51.37                                                             & 55.37                                                      \\ \hline
\multirow{4}{*}{PRMI}        & \multicolumn{5}{c|}{\multirow{2}{*}{No}}  & No              & 99.05                                                          & 97.90                                                             & \textbf{99.18}                                                   & 96.90                                                             & 98.25                                                      \\ \cline{7-12} 
                             & \multicolumn{5}{c|}{}                     & Yes             & 97.90                                                          & 95.98                                                             & 98.72                                                            & 97.72                                                             & \textbf{98.83}                                             \\ \cline{2-12} 
                             & \multicolumn{5}{c|}{\multirow{2}{*}{Yes}} & No              & 99.20                                                          & 97.95                                                             & \textbf{99.47}                                                   & 98.10                                                             & 99.17                                                      \\ \cline{7-12} 
                             & \multicolumn{5}{c|}{}                     & Yes             & 98.37                                                          & 97.58                                                             & \textbf{99.35}                                                   & 97.87                                                             & 98.63                                                      \\ \hline
\end{tabular}
\end{table*}
\subsection{Encoder Experiments}
For the two hyperparameters, $\lambda$ and $d$, the performance of the DREN models depended more on the selection of the embedding dimension. Ideally, as the dimensionality of the feature space becomes larger, inter-class separability should increase and intra-class variations should decrease leading to improved performance (as shown in our experiments for the DREN models). However, our DREN models achieved statistically comparable performance to the baseline models without an encoder at significantly lower dimensions (except for DREN with ResNet50 for DTD).  Our method is also stable to the selection of $\lambda$ except for when only the $L_{div}$ is considered ($\lambda$=1). In this case, the model is an unsupervised approach and will only learn embeddings that minimize the divergence between higher and lower dimension. In our implementation, the error was not backpropagated through the output layer once $\lambda=1$. In this instance, the output layer will not be updated, leading to poor classification performance. We also analyze the convergence of the proposed method across the various hyperparameters in the supplemental material (Section \ref{sec:supp}).

In Table \ref{tab:KTH_DREN}, the performance of the DREN models with and without the histogram layers for KTH-TIPS-2b were comparable. KTH-TIPS-2b primarily consisted of images collected in a controlled environment \cite{liu2019bow}. As a result, the local, texture features extracted by the histogram layers may not add as much information for the classification. However, the DREN models with the histogram layers performed better on DTD across the different combinations of hyperparameters as shown in Table \ref{tab:DTD_DREN}. DTD contained images collected ``in-the-wild" and also has more classes (47) than KTH-TIPS-2b (11). The embeddings learned by the DREN histogram models retained useful information from the texture features to achieve comparable performance to the base CNN models.

\subsection{t-SNE Comparisons}

In Table \ref{tab:KNN}, the test performance for the $K$-NN classifiers showed the utility of the DREN embeddings for not only visualization but also classification. Our DREN models also have an advantage over t-SNE in that new images can be embedded without a separate out-of-sample approach. In Figure \ref{fig:Embeddings}, the t-SNE embedding of the test points are not as visually compact and separable as the DREN models. The test samples embeddings produced from the DREN models also seem to have ``angular" feature distributions. As noted in other works \cite{choi2020amc}, models trained using cross-entropy appear to learn these unique, intrinsic feature coordinates in comparison to the t-SNE embeddings that are learned by only using KL divergence. For the DREN histogram models, the ``angular" embeddings may also be produced as a result of normalization as the features are also centered around $0$. The features captured by the histogram layer binning function are naturally between $0$ and $1$ while the features from the global average pooling layer are followed by batch normalization \cite{peeples2020histogram}. 

\subsection{Backbone Experiments}
The proposed DREN model improved the average test classification accuracy across various backbone architectures for the baseline and histogram layer models as shown in Table {\ref{tab:Architectures}}. We also observe that the features extracted from the histogram layer also led to the best performance when coupled with the divergence regulation term for the {KTH and DTD datasets. The structural changes of the roots are the most prominent features in the PRMI dataset, so the statistical information of the histogram-based models did not impact performance much in comparison to the base model.} The best model across all datasets was the DenseNet model. This model has a unique feature to encourage feature reuse {\cite{Huang2017DenseNets}} and this was more effective with the divergence regulation of the DREN model. As shown from the experiments, the DREN model is also fairly robust to the CNN backbone architecture.

\section{Conclusion}
In this work, we presented DREN models for joint dimensionality reduction and classification. The proposed approach learns discriminative features at lower dimensions that can be used for different tasks. The approach has several advantages and is a general framework that can use various classification/divergence measures and deep learning models. 

\section*{Acknowledgment}
This material is based upon work supported by the National Science Foundation Graduate Research Fellowship under Grant No. DGE-1842473 and by the Office of Naval Research grant N00014-16-1-2323. The views and opinions of authors expressed herein do not necessarily state or reflect those of the United States Government or any agency thereof.

\printbibliography

@article{shi2022remote,
  title={Remote Sensing Scene Image Classification Based on Self-Compensating Convolution Neural Network},
  author={Shi, Cuiping and Zhang, Xinlei and Sun, Jingwei and Wang, Liguo},
  journal={Remote Sensing},
  volume={14},
  number={3},
  pages={545},
  year={2022},
  publisher={Multidisciplinary Digital Publishing Institute}
}

@inproceedings{xu2022prmi,
  title={PRMI: A Dataset of Minirhizotron Images for Diverse Plant Root Study},
  author={Weihuang Xu and Guohao Yu and Yiming Cui and Romain Gloaguen and Alina Zare and Jason Bonnette 
      and Joel Reyes-Cabrera and Ashish Rajurkar and Diane Rowland and Roser Matamala and Julie D. Jastrow 
      and Thomas E. Juenger and Felix B. Fritschi},
  booktitle={Proceedings of the AAAI Conference on Artificial Intelligence AI for Agriculture and Food Systems (AIAFS) Workshops},
  pages={838--839},
  year={2022}
}

@article{VanDerMaaten2008Visualizing,
  title={Visualizing data using t-SNE},
  author={Maaten, Laurens van der and Hinton, Geoffrey},
  journal={Journal of machine learning research},
  volume={9},
  number={Nov},
  pages={2579--2605},
  year={2008}
}

@article{VanDerMaaten2009DRReview,
	author = {L. van der Maaten and E. Postma and H. Herik},
	year = {2007},
	month = {01},
	pages = {},
	title = {Dimensionality Reduction: A Comparative Review},
	volume = {10},
	journal = {Journal of Machine Learning Research - JMLR}
}

@article{Bengio2014RepLearningReview,
	author    = {Y. Bengio and A. C. Courville and P. Vincent},
	title     = {Unsupervised Feature Learning and Deep Learning: {A} Review and New
	Perspectives},
	journal   = {CoRR},
	volume    = {abs/1206.5538},
	year      = {2012},
	url       = {http://arxiv.org/abs/1206.5538},
	archivePrefix = {arXiv},
	eprint    = {1206.5538}
}

@phdthesis{Thorstensen2009ManifoldThesis,
	author = {N. Thorstensen},
	title = {Manifold learning and applications to shape and image processing},
	school = {Ecole Nationale des Ponts et Chaussees},
	year = {2009},
	address = {Paris, France},
	month = {Nov.}
}

@article{Roweis2000LLE,
	author = {S. T. Roweis and L. K. Saul},
	title = {Nonlinear Dimensionality Reduction by Locally Linear Embedding},
	volume = {290},
	number = {5500},
	pages = {2323--2326},
	year = {2000},
	doi = {10.1126/science.290.5500.2323},
	publisher = {American Association for the Advancement of Science},
	issn = {0036-8075},
	journal = {Science}
}

@article{Vural2018StudySupervisedManifoldLearning,
	author    = {E. Vural and C. Guillemot},
	title     = {A study of the classification of low-dimensional data with supervised manifold learning},
	journal   = {CoRR},
	volume    = {abs/1507.05880},
	year      = {2018},
	url       = {http://arxiv.org/abs/1507.05880},
	archivePrefix = {arXiv},
	eprint    = {1507.05880}
}

@Article{Chao2019RecentAdvancesSupervisedDimRed,
	AUTHOR = {G. Chao and Y. Luo and W. Ding},
	TITLE = {Recent Advances in Supervised Dimension Reduction: A Survey},
	JOURNAL = {Machine Learning and Knowledge Extraction},
	VOLUME = {1},
	YEAR = {2019},
	NUMBER = {1},
	PAGES = {341--358},
	ISSN = {2504-4990},
	DOI = {10.3390/make1010020}
}

@ARTICLE{Rish2008SupDimRedGLM,
	author = {I. Rish and G. Grabarnik and G. A. Cecchi and F. Pereira and G. J. Gordon},
	year = {2008},
	month = {01},
	pages = {832-839},
	title = {Closed-form supervised dimensionality reduction with generalized linear models},
	doi = {10.1145/1390156.1390261},
}

@article{Gorban2019BlessingofDimensionality,
	author = {Gorban, A. N.  and Tyukin, I. Y. },
	title = {Blessing of dimensionality: mathematical foundations of the statistical physics of data},
	journal = {Philosophical Transactions of the Royal Society A: Mathematical, Physical and Engineering Sciences},
	volume = {376},
	number = {2118},
	pages = {20170237},
	year = {2018},
	doi = {10.1098/rsta.2017.0237},
}

@article{Huang2019UnderstandingGeneralization,
	author    = {W. Ronny Huang and
	Zeyad Emam and
	Micah Goldblum and
	Liam Fowl and
	Justin K. Terry and
	Furong Huang and
	Tom Goldstein},
	title     = {Understanding Generalization through Visualizations},
	journal   = {CoRR},
	volume    = {abs/1906.03291},
	year      = {2019},
	url       = {http://arxiv.org/abs/1906.03291},
	archivePrefix = {arXiv},
	eprint    = {1906.03291}
}

@INPROCEEDINGS{Huang2017DenseNets,
  author={G. {Huang} and Z. {Liu} and L. {Van Der Maaten} and K. Q. {Weinberger}},
  booktitle={2017 IEEE Conference on Computer Vision and Pattern Recognition (CVPR)}, 
  title={Densely Connected Convolutional Networks}, 
  year={2017},
  volume={},
  number={},
  pages={2261-2269},
  doi={10.1109/CVPR.2017.243}
  }

@article{peeples2020histogram,
  title={Histogram layers for texture analysis},
  author={Peeples, Joshua and Xu, Weihuang and Zare, Alina},
  journal={IEEE Transactions on Artificial Intelligence},
  year={2021},
  publisher={IEEE}
}

@InProceedings{cimpoi14describing,
	Author    = {M. Cimpoi and S. Maji and I. Kokkinos and S. Mohamed and A. Vedaldi},
	Title     = {Describing Textures in the Wild},
	Booktitle = {Proceedings of the {IEEE} Conf. on Computer Vision and Pattern Recognition ({CVPR})},
	Year      = {2014}}

@article{KTHTIPS_Databases,
  title={The kth-tips2 database},
  author={Mallikarjuna, P and Targhi, Alireza Tavakoli and Fritz, Mario and Hayman, Eric and Caputo, Barbara and Eklundh, Jan-Olof},
  journal={KTH Royal Institute of Technology},
  year={2006}
}

@article{hiremath2007content,
  title={Content Based Image Retrieval based on Color, Texture and Shape features using Image and its complement},
  author={Hiremath, PS and Pujari, Jagadeesh},
  journal={International Journal of Computer Science and Security},
  volume={1},
  number={4},
  pages={25--35},
  year={2007}
}

@article{liu2019bow,
  title={From BoW to CNN: Two decades of texture representation for texture classification},
  author={Liu, Li and Chen, Jie and Fieguth, Paul and Zhao, Guoying and Chellappa, Rama and Pietik{\"a}inen, Matti},
  journal={International Journal of Computer Vision},
  volume={127},
  number={1},
  pages={74--109},
  year={2019},
  publisher={Springer}
}

@inproceedings{choi2020amc,
  title={AMC-Loss: Angular Margin Contrastive Loss for Improved Explainability in Image Classification},
  author={Choi, Hongjun and Som, Anirudh and Turaga, Pavan},
  booktitle={Proceedings of the IEEE/CVF Conference on Computer Vision and Pattern Recognition Workshops},
  pages={838--839},
  year={2020}
}

@article{Hong2019LearnableManifoldAlignment,
	title = {Learnable manifold alignment (LeMA): A semi-supervised cross-modality learning framework for land cover and land use classification},
	author = {D. Hong and N. Yokoya and N. Ge and J. Chanussot and X. X. Zhu},
	journal = {ISPRS Journal of Photogrammetry and Remote Sensing},
	volume = {147},
	pages = {193 -- 205},
	year = {2019},
	issn = {0924-2716},
}

@article{Hong2018CommonSubspaceLearningHSI,
	author    = {D. Hong and N. Yokoya and J. Chanussot and	X. X. Zhu},
	title     = {CoSpace: Common Subspace Learning from Hyperspectral-Multispectral Correspondences},
	journal   = {CoRR},
	volume    = {abs/1812.11501},
	year      = {2018},
	url       = {http://arxiv.org/abs/1812.11501},
	archivePrefix = {arXiv},
	eprint    = {1812.11501},
}

@ARTICLE{Duan2021DimRedHSI,  
author={Duan, Yule and Huang, Hong and Tang, Yuxiao},  
journal={IEEE Transactions on Geoscience and Remote Sensing},   
title={Local Constraint-Based Sparse Manifold Hypergraph Learning for Dimensionality Reduction of Hyperspectral Image},   
year={2021},  
volume={59},  
number={1},  
pages={613-628},  
doi={10.1109/TGRS.2020.2995709}
}

@ARTICLE{Luo2021DimRedHSI,  
author={Luo, Fulin and Zou, Zehua and Liu, Jiamin and Lin, Zhiping},  
journal={IEEE Transactions on Geoscience and Remote Sensing},   
title={Dimensionality reduction and classification of hyperspectral image via multi-structure unified discriminative embedding},   
year={2021},  
volume={},  
number={},  
pages={1-1},  
doi={10.1109/TGRS.2021.3128764}
}
\clearpage
\section{Supplemental Material} \label{sec:supp}
\begin{figure*}[htb]
    \centering
	\begin{subfigure}{.195\textwidth}{
			\includegraphics[width=\textwidth]{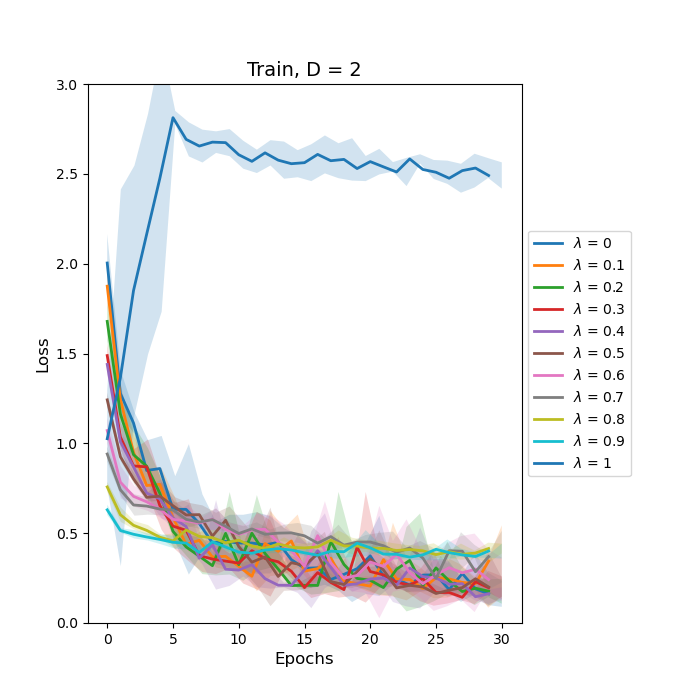}
			\caption{Train, $d = 2$}
			\label{fig:KTH_Base_Train_2D}
		}
	\end{subfigure}
	\centering
	\begin{subfigure}{.195\textwidth}{
	\includegraphics[width=\textwidth]{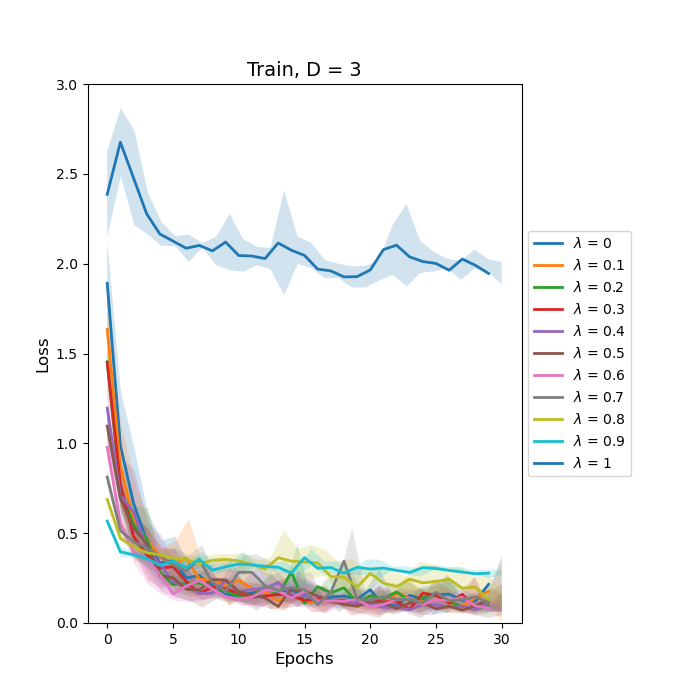}
			\caption{Train, $d = 3$}
			\label{fig:KTH_Base_Train_3D}
		}
	\end{subfigure}
	\centering
	\begin{subfigure}{.195\textwidth}{
	\includegraphics[width=\textwidth]{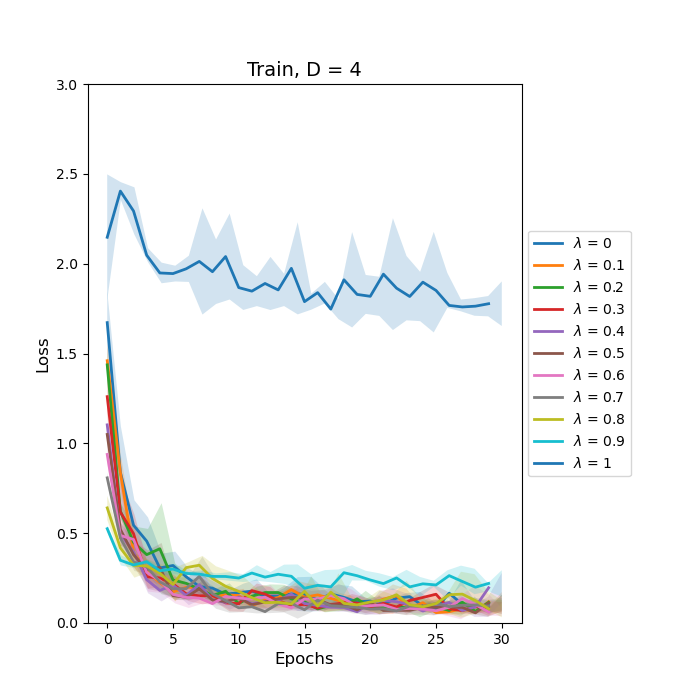}
			\caption{Train, $d = 4$}
			\label{fig:KTH_Base_Train_4D}
		}
	\end{subfigure} 
	\centering
	\begin{subfigure}{.195\textwidth}{
	\includegraphics[width=\textwidth]{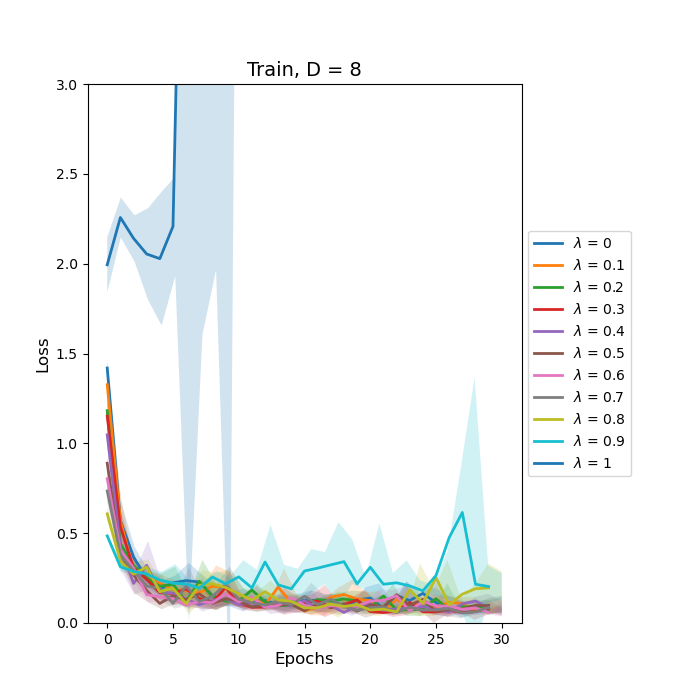}
			\caption{Train, $d = 8$}
			\label{fig:KTH_Base_Train_8D}
		}
	\end{subfigure}
	\centering
		\begin{subfigure}{.195\textwidth}{
	\includegraphics[width=\textwidth]{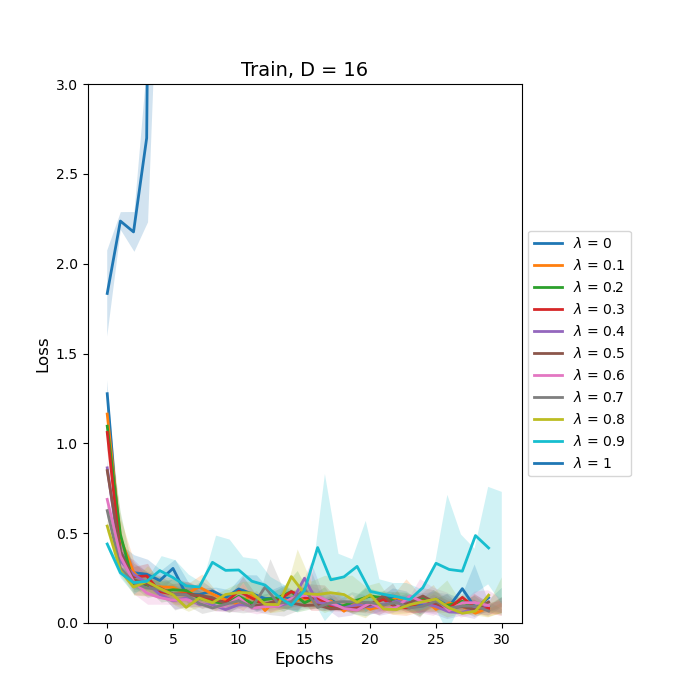}
			\caption{Train, $d = 16$}
			\label{fig:KTH_Base_Train_16D}
		}
	\end{subfigure}
	
    \centering
	\begin{subfigure}{.195\textwidth}{
			\includegraphics[width=\textwidth]{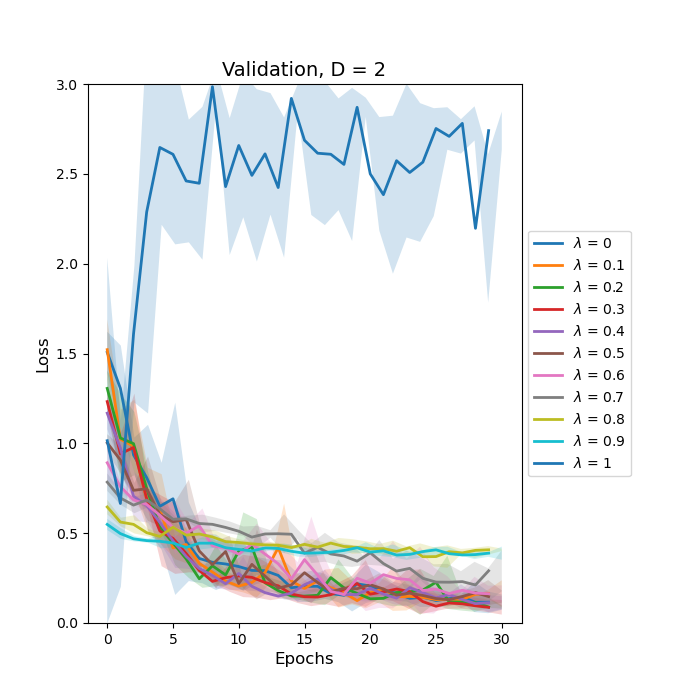}
			\caption{Validation, $d = 2$}
			\label{fig:KTH_Base_Val_2D}
		}
	\end{subfigure}
	\centering
	\begin{subfigure}{.195\textwidth}{
	\includegraphics[width=\textwidth]{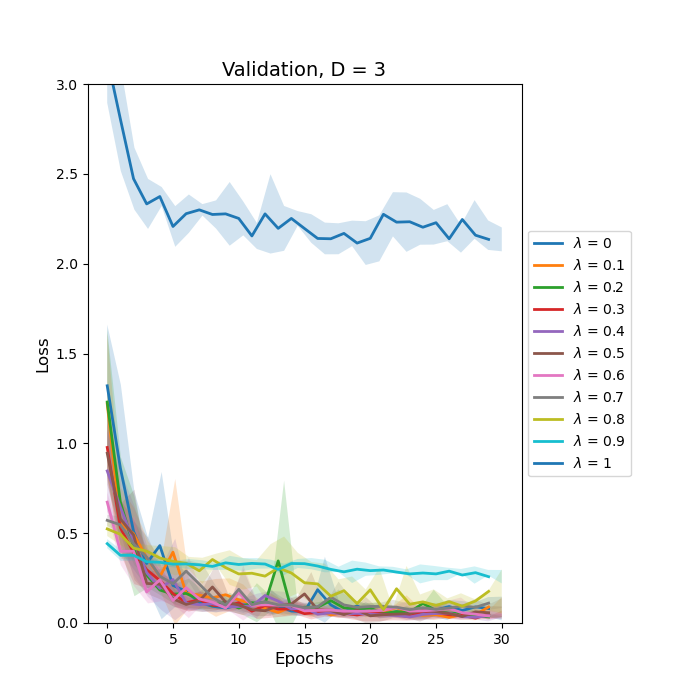}
			\caption{Validation, $d = 3$}
			\label{fig:KTH_Base_Val_3D}
		}
	\end{subfigure}
	\centering
	\begin{subfigure}{.195\textwidth}{
	\includegraphics[width=\textwidth]{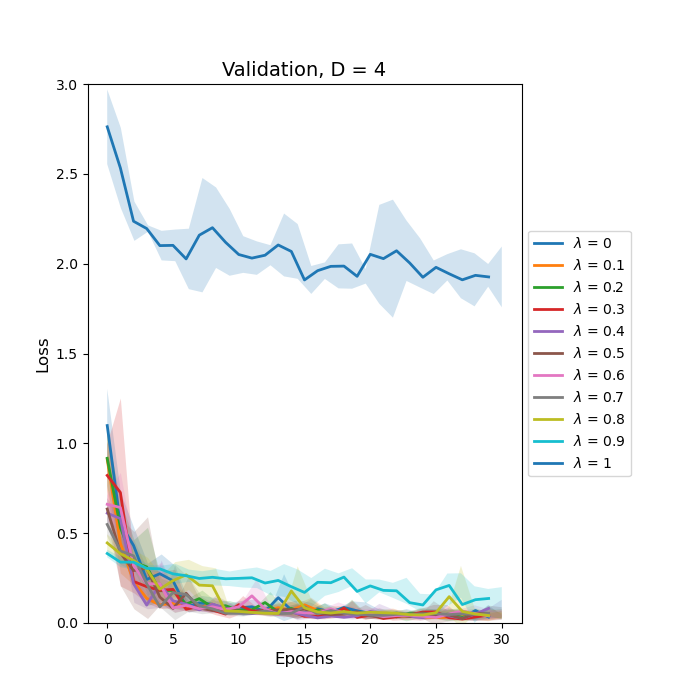}
			\caption{Validation, $d = 4$}
			\label{fig:KTH_Base_Val_4D}
		}
	\end{subfigure} 
	\centering
	\begin{subfigure}{.195\textwidth}{
	\includegraphics[width=\textwidth]{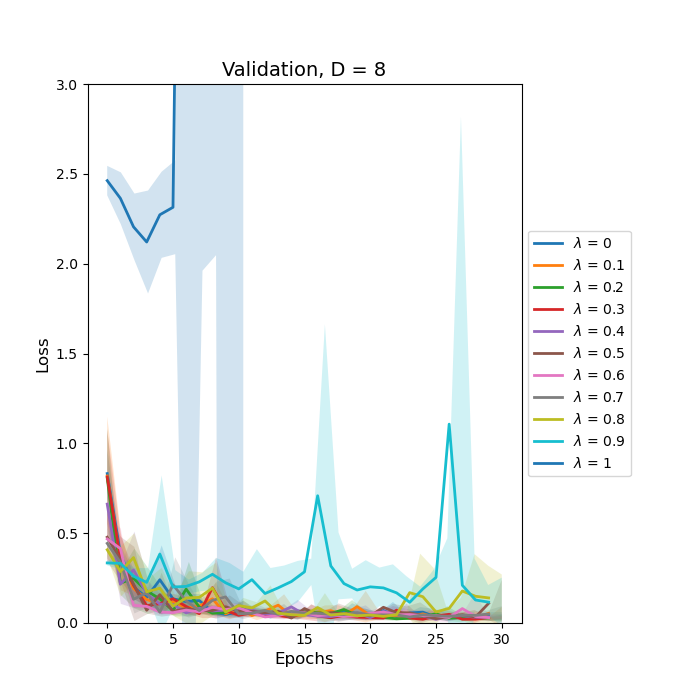}
			\caption{Validation, $d = 8$}
			\label{fig:KTH_Base_Val_8D}
		}
	\end{subfigure}
	\centering
		\begin{subfigure}{.195\textwidth}{
	\includegraphics[width=\textwidth]{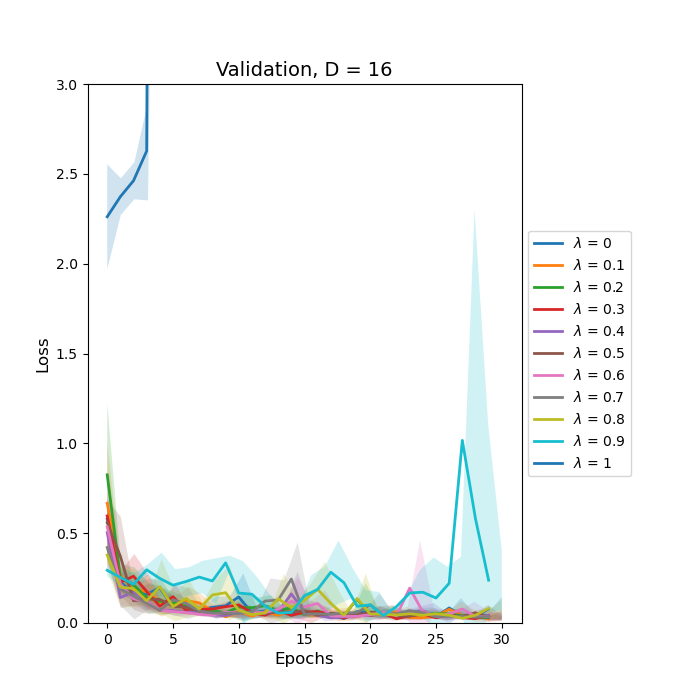}
			\caption{Validation, $d = 16$}
			\label{fig:KTH_Base_Val_16D}
		}
	\end{subfigure}

	\caption{Learning curves for ResNet18 model on KTH-TIPS-2b dataset. The top and bottom row show the average loss with standard deviation for the training and validation data respectively. Each column shows the results for a value of $d$: 2, 3, 4, 8, and 16. }
	\label{fig:KTH_Base} 
\end{figure*}

\begin{figure*}[htb]
    \centering
	\begin{subfigure}{.195\textwidth}{
			\includegraphics[width=\textwidth]{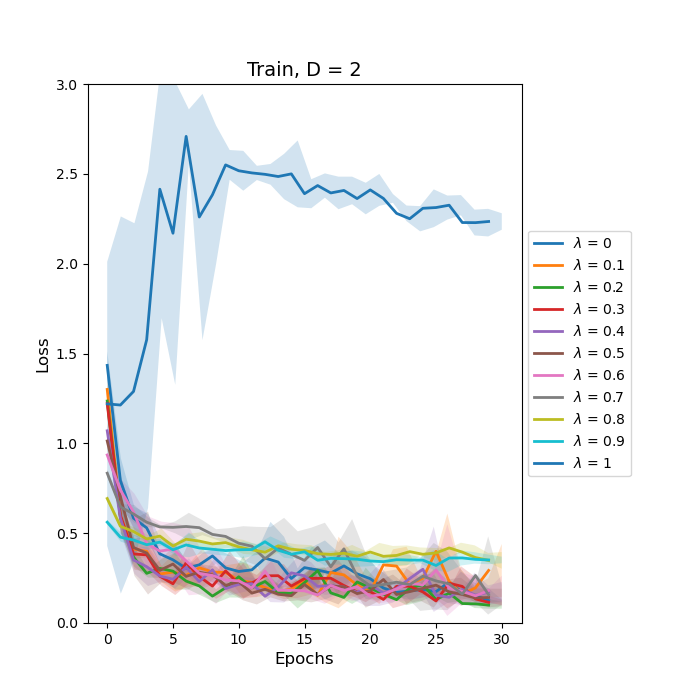}
			\caption{Train, $d = 2$}
			\label{fig:KTH_Hist_Train_2D}
		}
	\end{subfigure}
	\centering
	\begin{subfigure}{.195\textwidth}{
	\includegraphics[width=\textwidth]{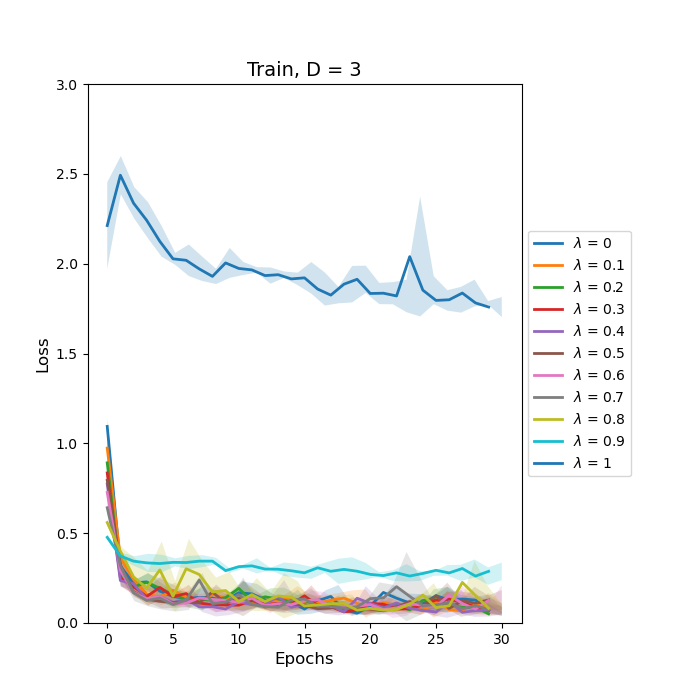}
			\caption{Train, $d = 3$}
			\label{fig:KTH_Hist_Train_3D}
		}
	\end{subfigure}
	\centering
	\begin{subfigure}{.195\textwidth}{
	\includegraphics[width=\textwidth]{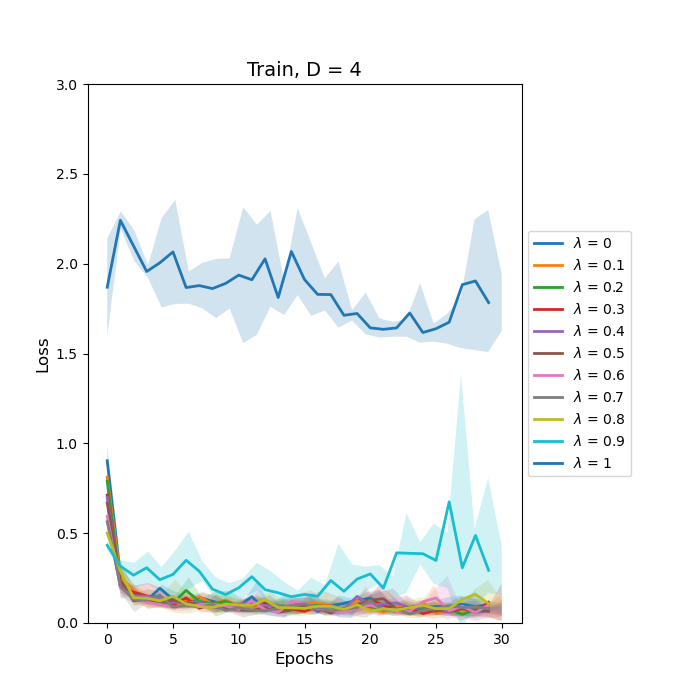}
			\caption{Train, $d = 4$}
			\label{fig:KTH_Hist_Train_4D}
		}
	\end{subfigure} 
	\centering
	\begin{subfigure}{.195\textwidth}{
	\includegraphics[width=\textwidth]{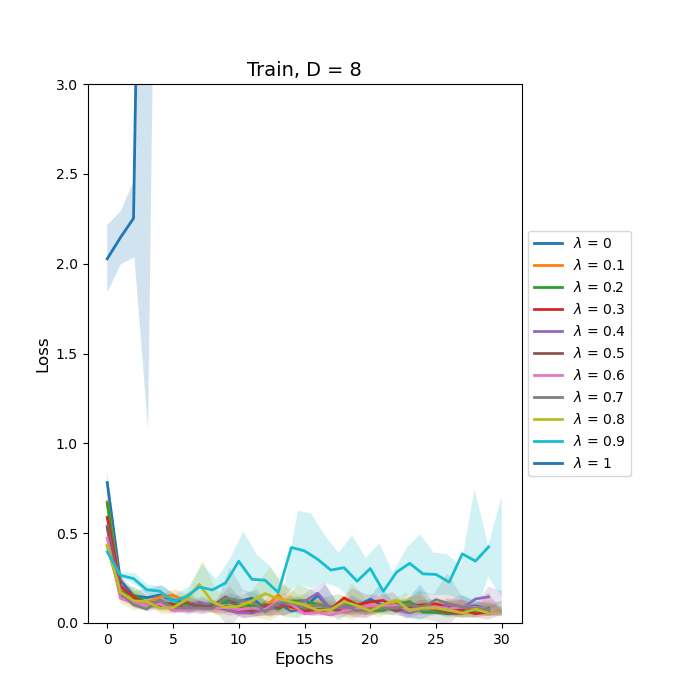}
			\caption{Train, $d = 8$}
			\label{fig:KTH_Hist_Train_8D}
		}
	\end{subfigure}
	\centering
		\begin{subfigure}{.195\textwidth}{
	\includegraphics[width=\textwidth]{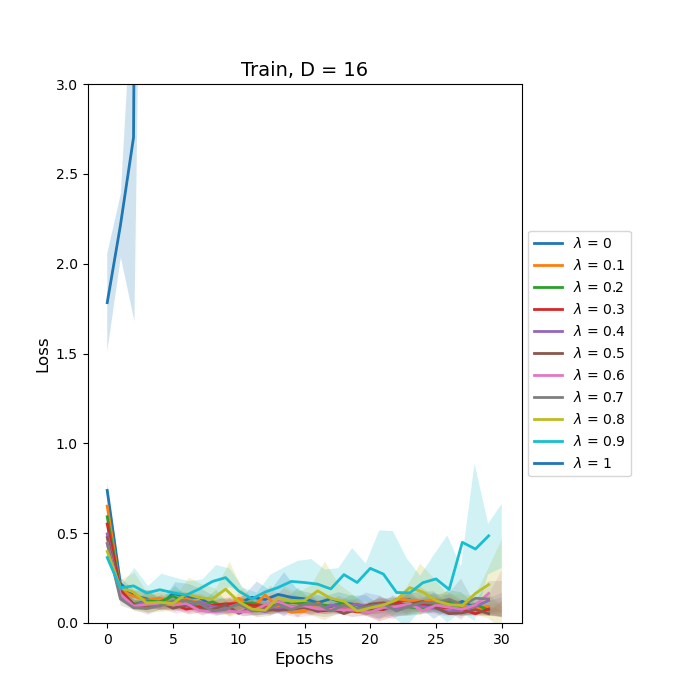}
			\caption{Train, $d = 16$}
			\label{fig:KTH_Hist_Train_16D}
		}
	\end{subfigure}
	
    \centering
	\begin{subfigure}{.195\textwidth}{
			\includegraphics[width=\textwidth]{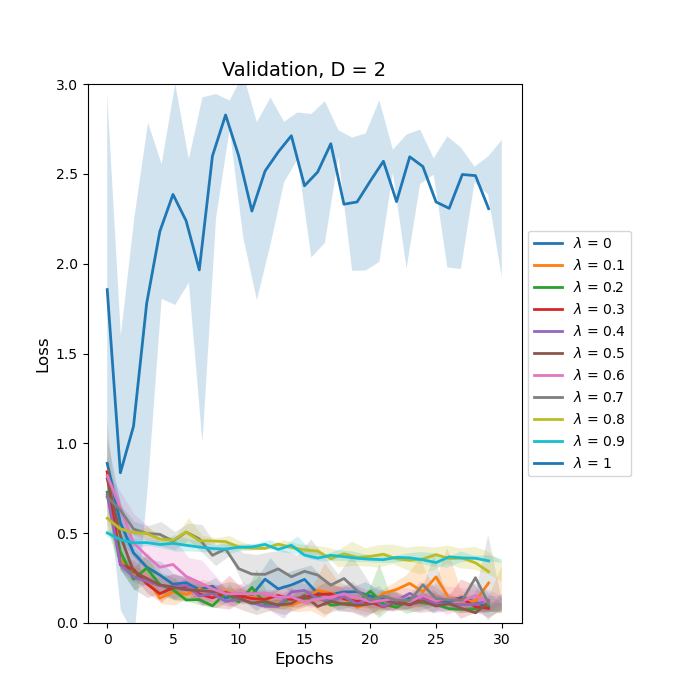}
			\caption{Validation, $d = 2$}
			\label{fig:KTH_Hist_Val_2D}
		}
	\end{subfigure}
	\centering
	\begin{subfigure}{.195\textwidth}{
	\includegraphics[width=\textwidth]{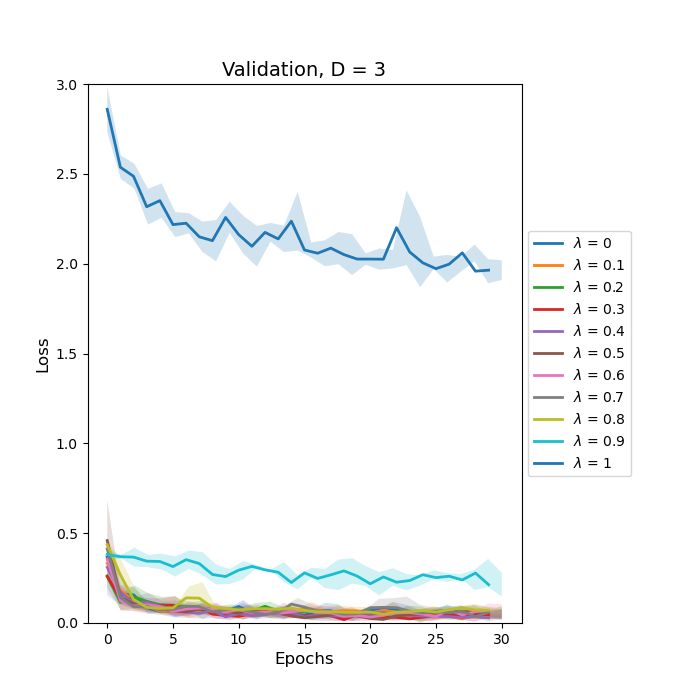}
			\caption{Validation, $d = 3$}
			\label{fig:KTH_Hist_Val_3D}
		}
	\end{subfigure}
	\centering
	\begin{subfigure}{.195\textwidth}{
	\includegraphics[width=\textwidth]{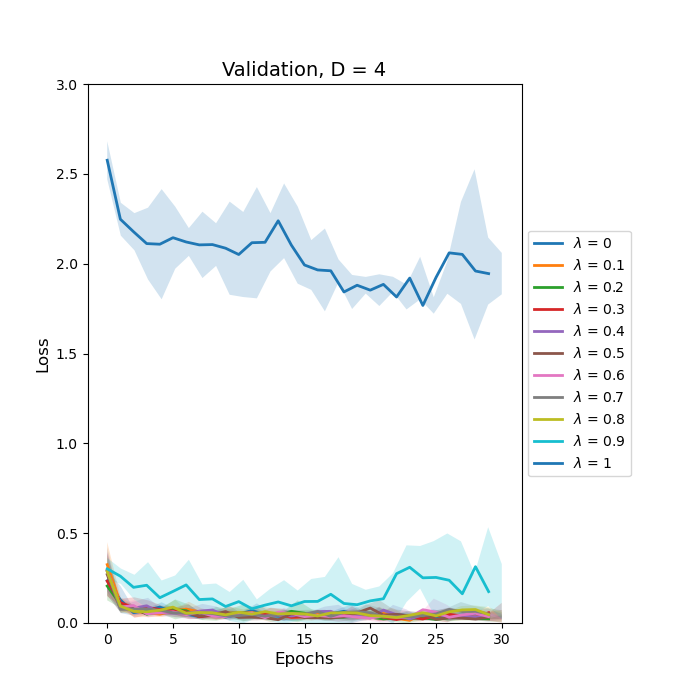}
			\caption{Validation, $d = 4$}
			\label{fig:KTH_Hist_Val_4D}
		}
	\end{subfigure} 
	\centering
	\begin{subfigure}{.195\textwidth}{
	\includegraphics[width=\textwidth]{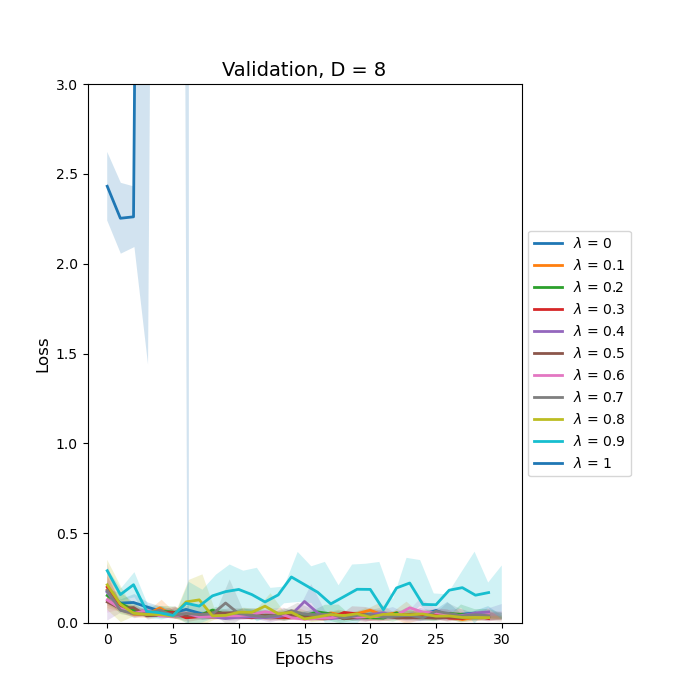}
			\caption{Validation, $d = 8$}
			\label{fig:KTH_Hist_Val_8D}
		}
	\end{subfigure}
	\centering
		\begin{subfigure}{.195\textwidth}{
	\includegraphics[width=\textwidth]{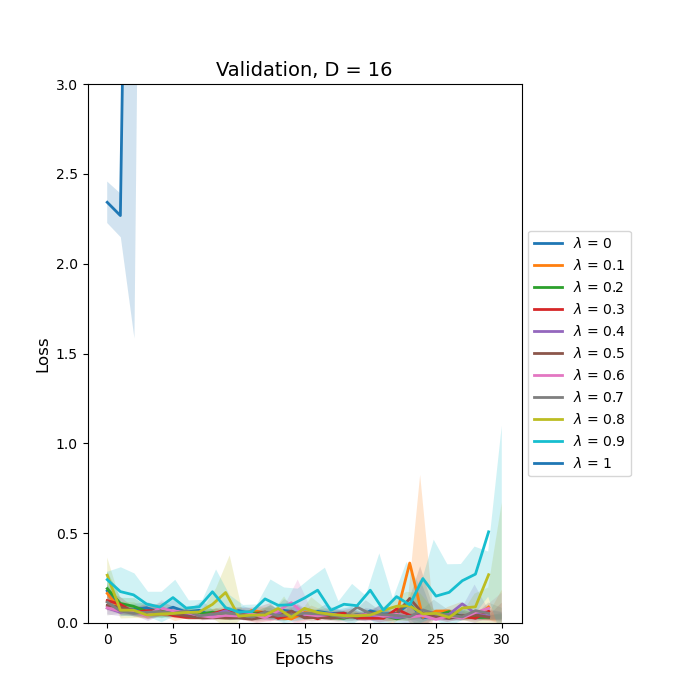}
			\caption{Validation, $d = 16$}
			\label{fig:KTH_Hist_Val_16D}
		}
	\end{subfigure}

	\caption{Learning curves for HistRes18 model on KTH-TIPS-2b dataset. The top and bottom row show the average loss with standard deviation for the training and validation data respectively. Each column shows the results for a value of $d$: 2, 3, 4, 8, and 16. }
	\label{fig:KTH_Hist} 
\end{figure*}

For the encoder experiments, we analyzed the learning curves of each model across the hyperparameters, $\lambda$ and $d$. The training and validation curves for ResNet18 and HistRes18 on the KTH-TIPS-2b dataset are shown in Figures \ref{fig:KTH_Base} and \ref{fig:KTH_Hist} respectively. We observe here that for most values of labmda, the model convergences as expected on the training and validation datasets. The only value of lambda that the model does not converge for is when $\lambda = 1$. In our implementation, the loss only consists of the divergence term and this loss is not backpropagated through the output layer. As shown in Figures \ref{fig:KTH_Base} and \ref{fig:KTH_Hist}, this serves as the reason why the loss did not trend lower as is the case for other values of $\lambda$. Another interesting observation is that as dimensionality ($d$) increased, the model converged at a faster rate. This result is intuitive as the model complexity (\textit{i.e.}, model parameters) increases, the algorithm can quickly learn (and possibly overfit) the training data.

\begin{figure*}[htb]
    \centering
	\begin{subfigure}{.195\textwidth}{
			\includegraphics[width=\textwidth]{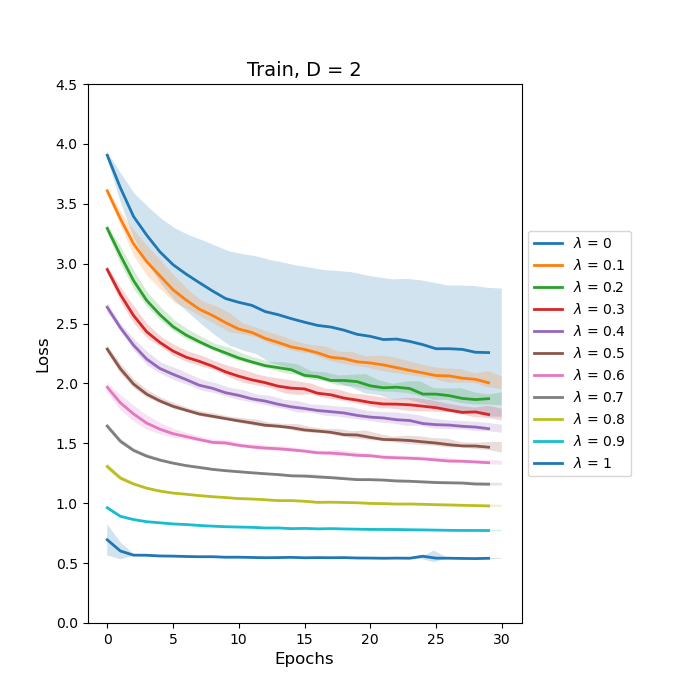}
			\caption{Train, $d = 2$}
			\label{fig:DTD_Base_Train_2D}
		}
	\end{subfigure}
	\centering
	\begin{subfigure}{.195\textwidth}{
	\includegraphics[width=\textwidth]{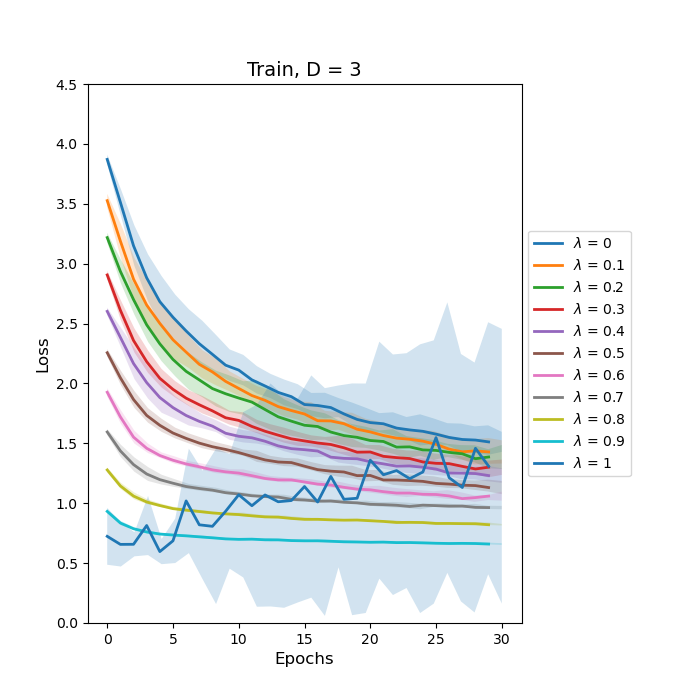}
			\caption{Train, $d = 3$}
			\label{fig:DTD_Base_Train_3D}
		}
	\end{subfigure}
	\centering
	\begin{subfigure}{.195\textwidth}{
	\includegraphics[width=\textwidth]{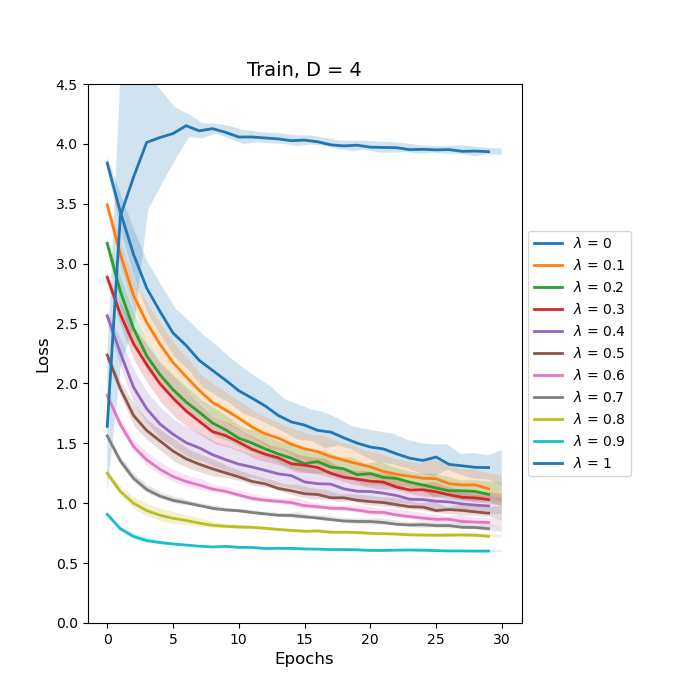}
			\caption{Train, $d = 4$}
			\label{fig:DTD_Base_Train_4D}
		}
	\end{subfigure} 
	\centering
	\begin{subfigure}{.195\textwidth}{
	\includegraphics[width=\textwidth]{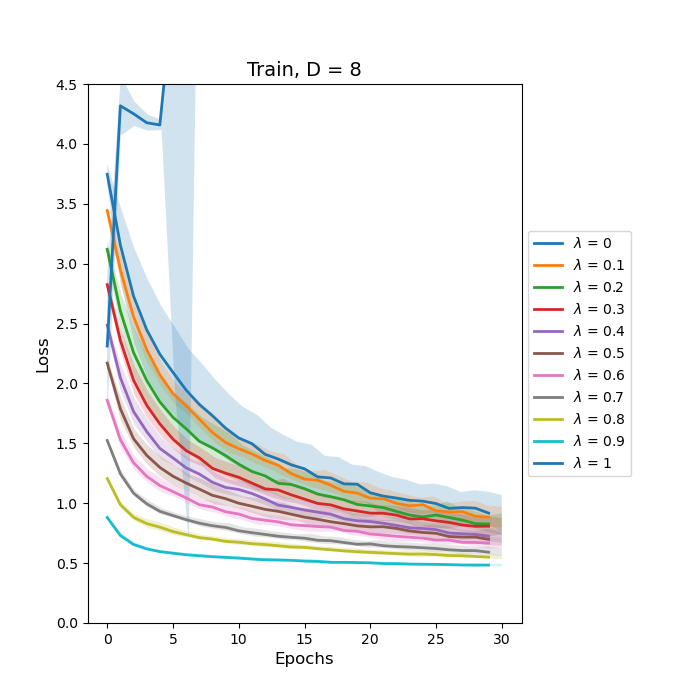}
			\caption{Train, $d = 8$}
			\label{fig:DTD_Base_Train_8D}
		}
	\end{subfigure}
	\centering
		\begin{subfigure}{.195\textwidth}{
	\includegraphics[width=\textwidth]{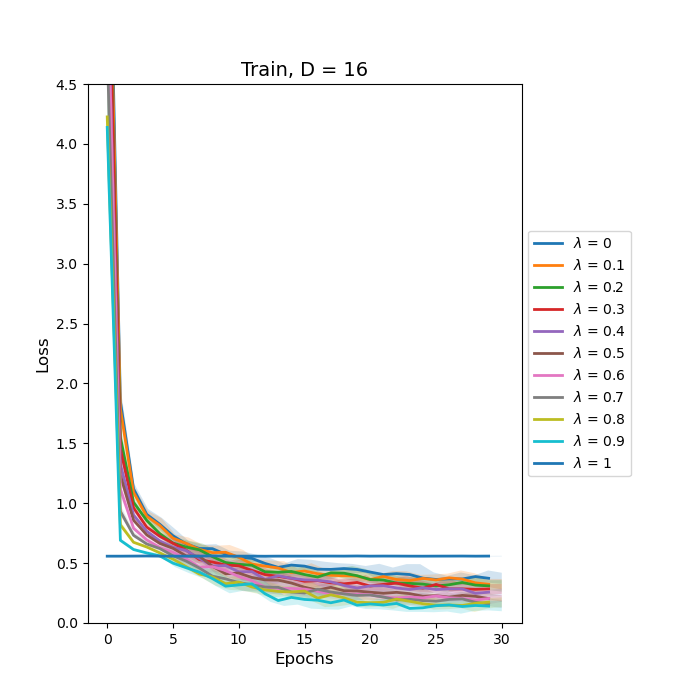}
			\caption{Train, $d = 16$}
			\label{fig:DTD_Base_Train_16D}
		}
	\end{subfigure}
	
    \centering
	\begin{subfigure}{.195\textwidth}{
			\includegraphics[width=\textwidth]{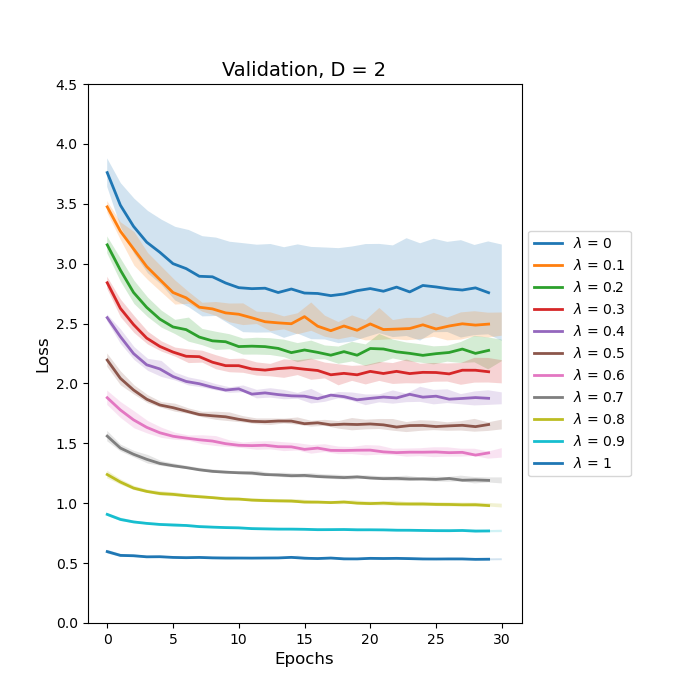}
			\caption{Validation, $d = 2$}
			\label{fig:DTD_Base_Val_2D}
		}
	\end{subfigure}
	\centering
	\begin{subfigure}{.195\textwidth}{
	\includegraphics[width=\textwidth]{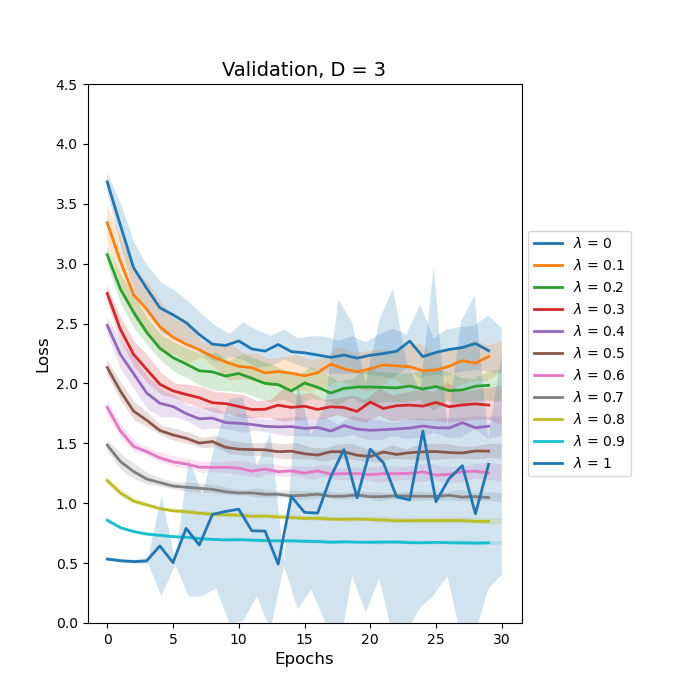}
			\caption{Validation, $d = 3$}
			\label{fig:DTD_Base_Val_3D}
		}
	\end{subfigure}
	\centering
	\begin{subfigure}{.195\textwidth}{
	\includegraphics[width=\textwidth]{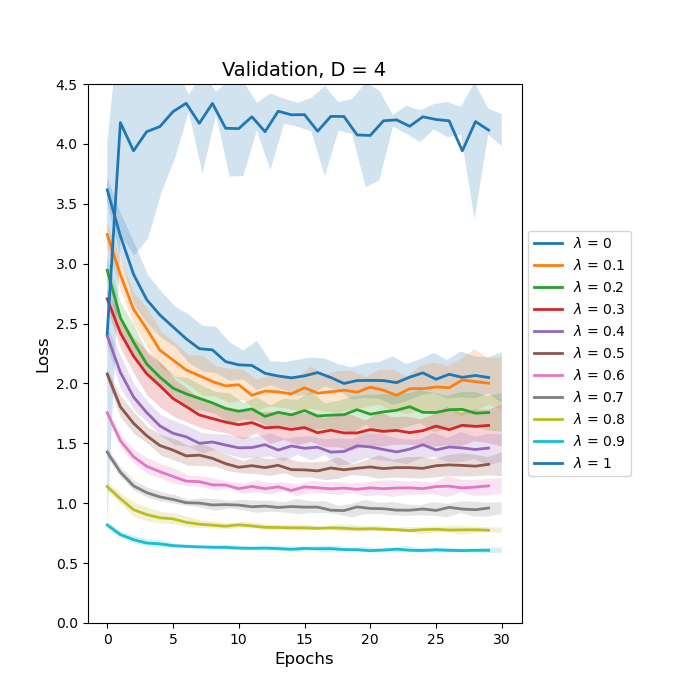}
			\caption{Validation, $d = 4$}
			\label{fig:DTD_Base_Val_4D}
		}
	\end{subfigure} 
	\centering
	\begin{subfigure}{.195\textwidth}{
	\includegraphics[width=\textwidth]{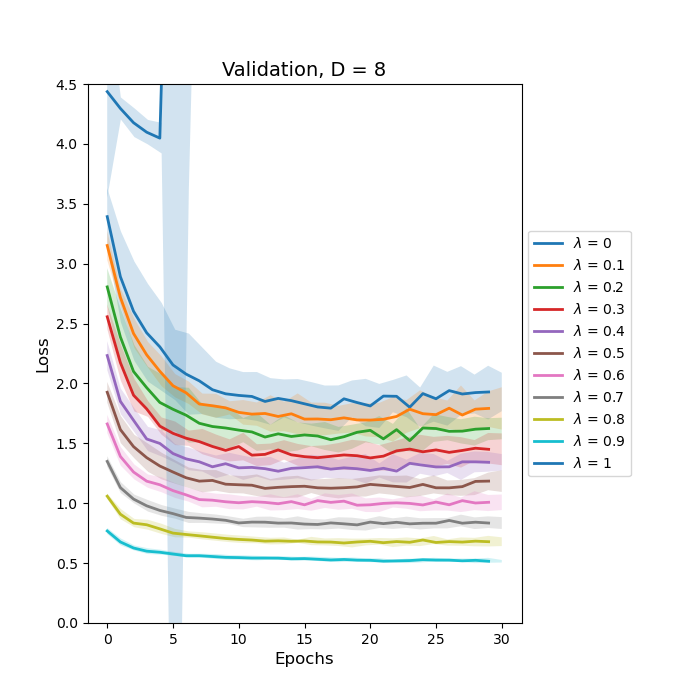}
			\caption{Validation, $d = 8$}
			\label{fig:DTD_Base_Val_8D}
		}
	\end{subfigure}
	\centering
		\begin{subfigure}{.195\textwidth}{
	\includegraphics[width=\textwidth]{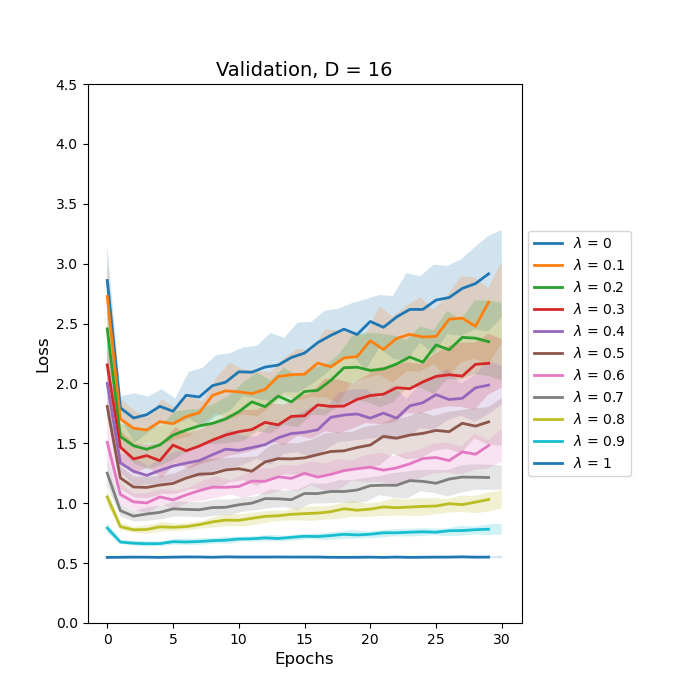}
			\caption{Validation, $d = 16$}
			\label{fig:DTD_Base_Val_16D}
		}
	\end{subfigure}

	\caption{Learning curves for ResNet50 model on DTD dataset. The top and bottom row show the average loss with standard deviation for the training and validation data respectively. Each column shows the results for a value of $d$: 2, 3, 4, 8, and 16. }
	\label{fig:DTD_Base} 
\end{figure*}

\begin{figure*}[htb]
    \centering
	\begin{subfigure}{.195\textwidth}{
			\includegraphics[width=\textwidth]{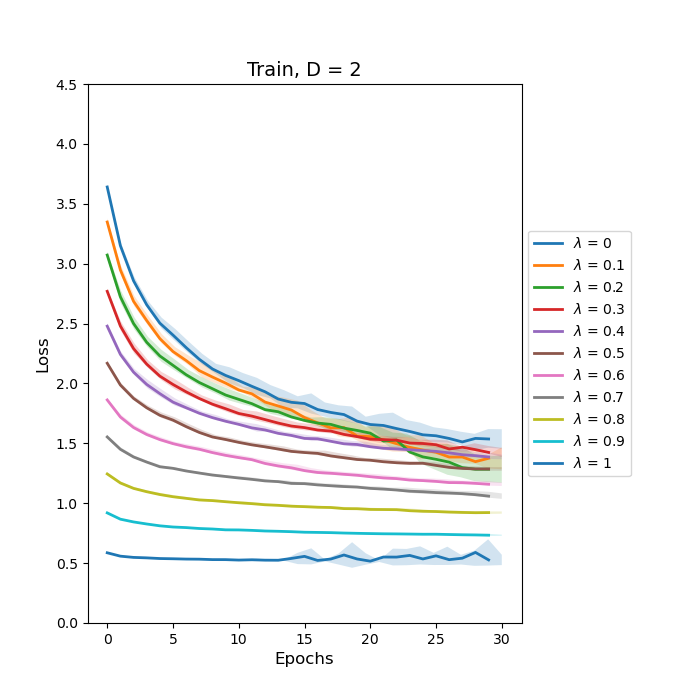}
			\caption{Train, $d = 2$}
			\label{fig:DTD_Hist_Train_2D}
		}
	\end{subfigure}
	\centering
	\begin{subfigure}{.195\textwidth}{
	\includegraphics[width=\textwidth]{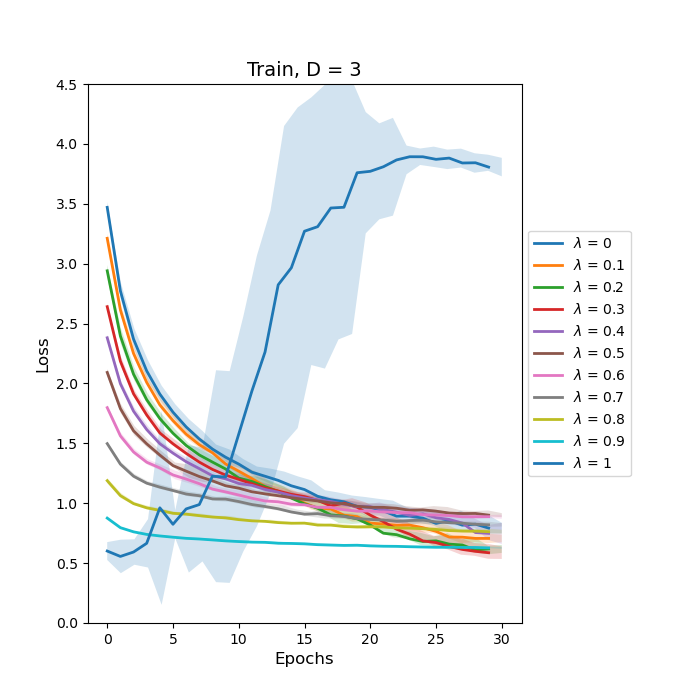}
			\caption{Train, $d = 3$}
			\label{fig:DTD_Hist_Train_3D}
		}
	\end{subfigure}
	\centering
	\begin{subfigure}{.195\textwidth}{
	\includegraphics[width=\textwidth]{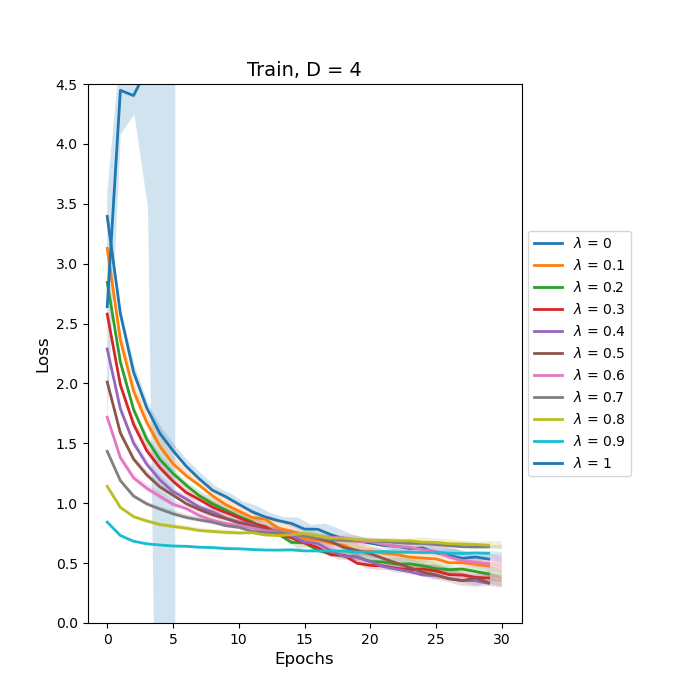}
			\caption{Train, $d = 4$}
			\label{fig:DTD_Hist_Train_4D}
		}
	\end{subfigure} 
	\centering
	\begin{subfigure}{.195\textwidth}{
	\includegraphics[width=\textwidth]{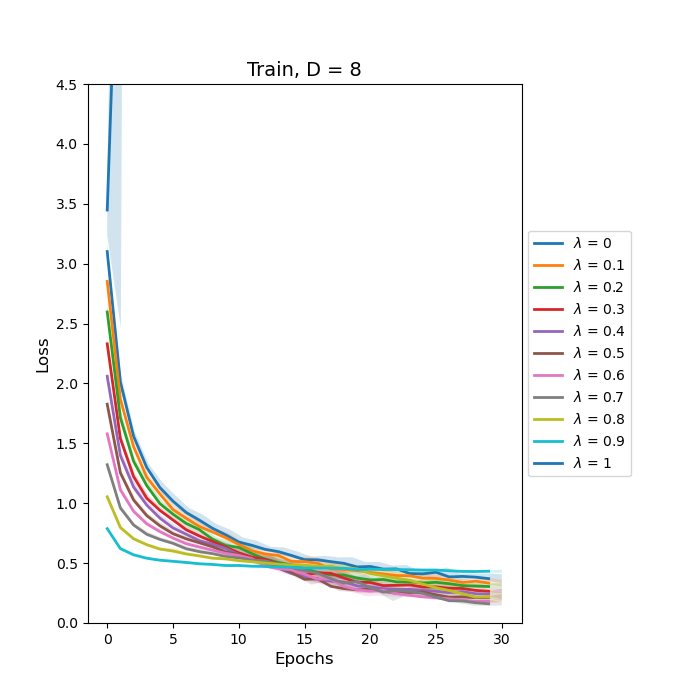}
			\caption{Train, $d = 8$}
			\label{fig:DTD_Hist_Train_8D}
		}
	\end{subfigure}
	\centering
		\begin{subfigure}{.195\textwidth}{
	\includegraphics[width=\textwidth]{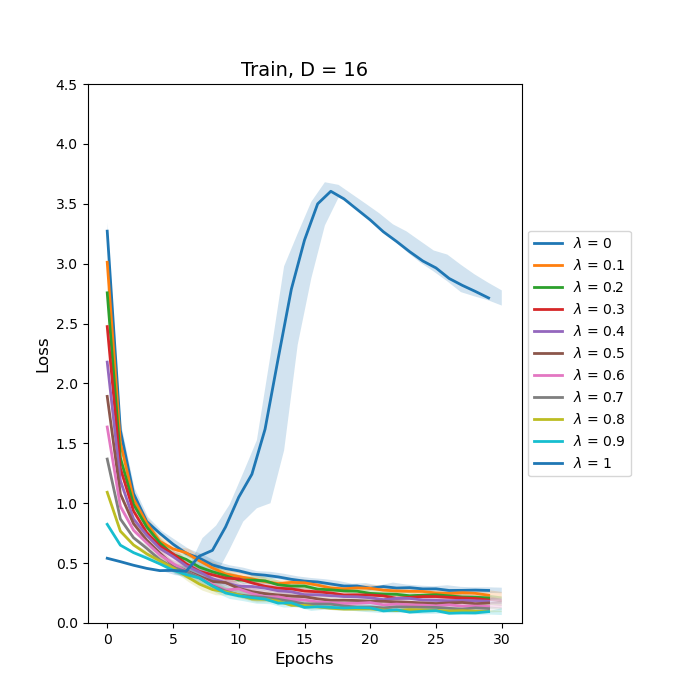}
			\caption{Train, $d = 16$}
			\label{fig:DTD_Hist_Train_16D}
		}
	\end{subfigure}
	
    \centering
	\begin{subfigure}{.195\textwidth}{
			\includegraphics[width=\textwidth]{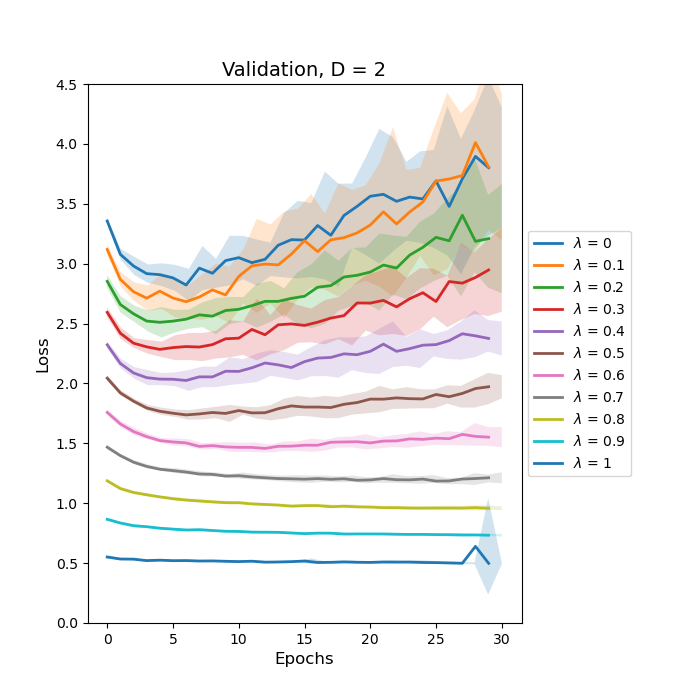}
			\caption{Validation, $d = 2$}
			\label{fig:DTD_Hist_Val_2D}
		}
	\end{subfigure}
	\centering
	\begin{subfigure}{.195\textwidth}{
	\includegraphics[width=\textwidth]{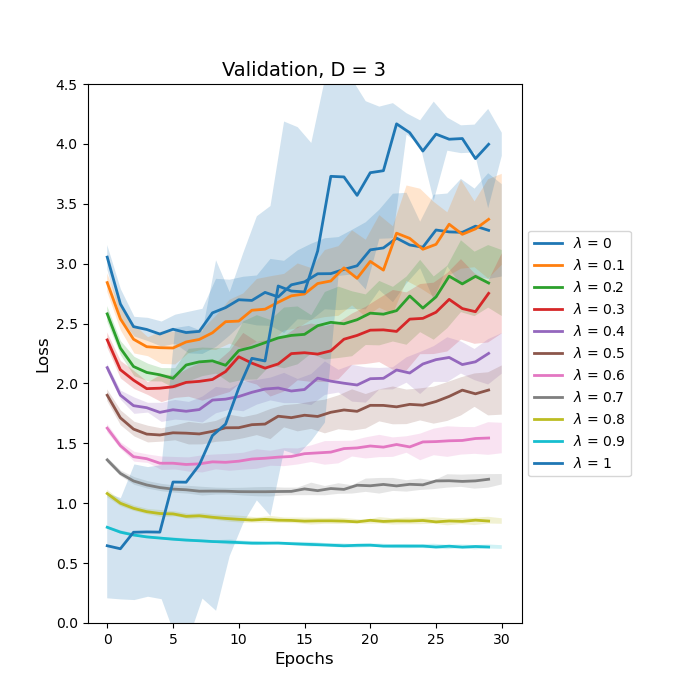}
			\caption{Validation, $d = 3$}
			\label{fig:DTD_Hist_Val_3D}
		}
	\end{subfigure}
	\centering
	\begin{subfigure}{.195\textwidth}{
	\includegraphics[width=\textwidth]{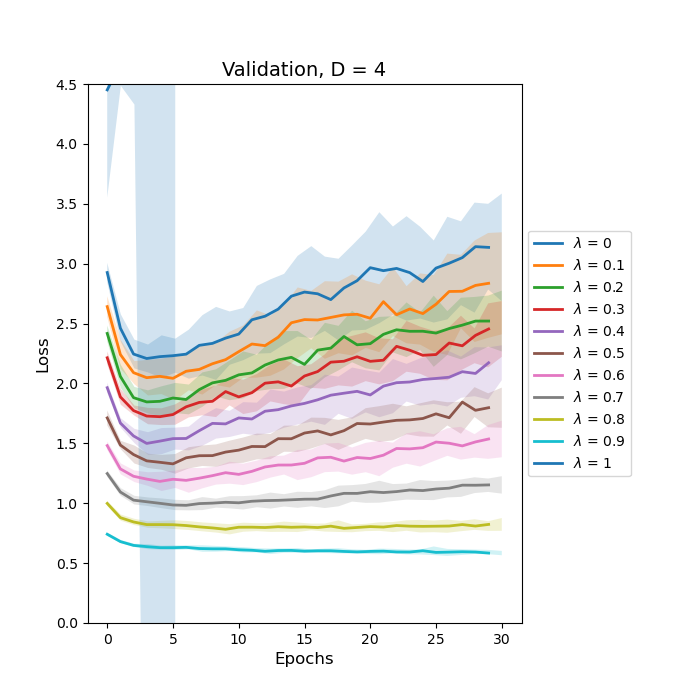}
			\caption{Validation, $d = 4$}
			\label{fig:DTD_Hist_Val_4D}
		}
	\end{subfigure} 
	\centering
	\begin{subfigure}{.195\textwidth}{
	\includegraphics[width=\textwidth]{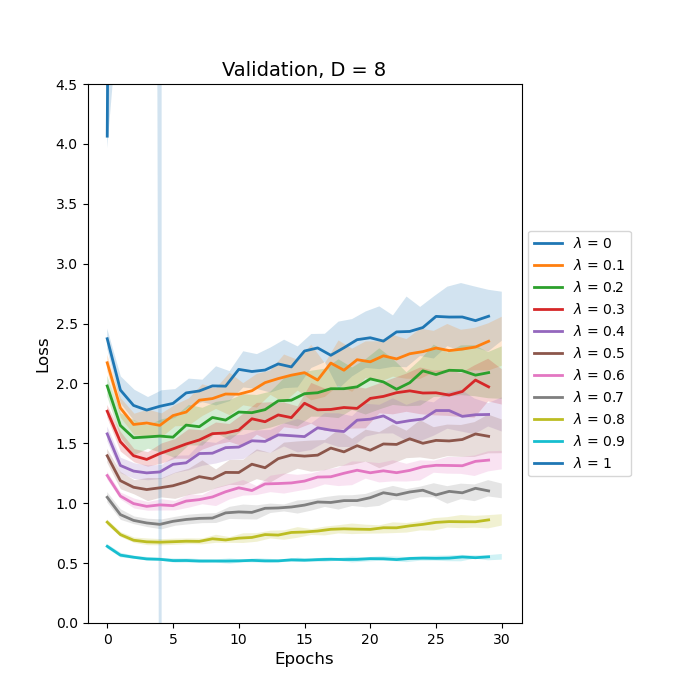}
			\caption{Validation, $d = 8$}
			\label{fig:DTD_Hist_Val_8D}
		}
	\end{subfigure}
	\centering
		\begin{subfigure}{.195\textwidth}{
	\includegraphics[width=\textwidth]{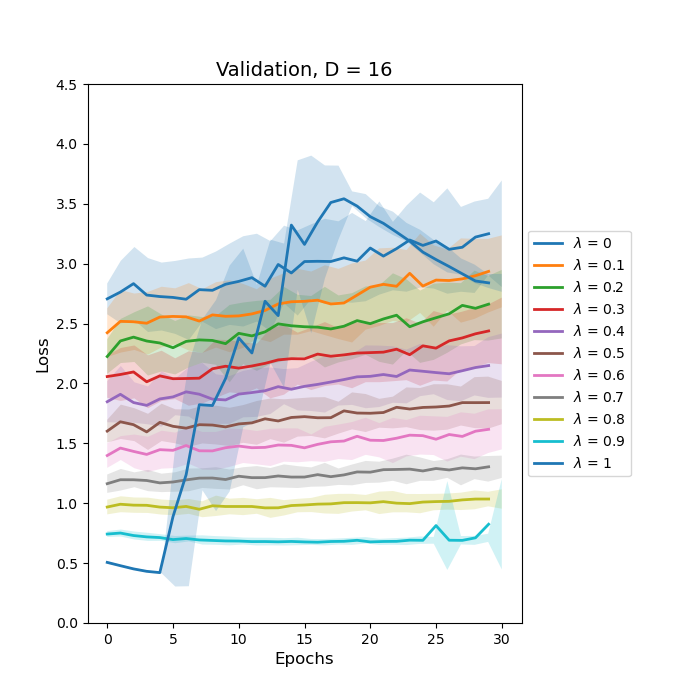}
			\caption{Validation, $d = 16$}
			\label{fig:DTD_Hist_Val_16D}
		}
	\end{subfigure}

	\caption{Learning curves for HistRes50 model on DTD dataset respectively. The top and bottom row show the average loss with standard deviation for the training and validation data. Each column shows the results for a value of $d$: 2, 3, 4, 8, and 16. }
	\label{fig:DTD_Hist} 
\end{figure*}

Similar observations are made with the results of the DTD dataset. The learning curves for the ResNet50 and HistRes50 model are shown in Figures \ref{fig:DTD_Base} and \ref{fig:DTD_Hist}. The DREN models converge for all values of $\lambda$ except for $\lambda = 1$ as in the case for the KTH-TIPS-2b dataset. As $d$ increases for the DREN ResNet50 and HistRes50 models, we observe that validation curves begin to increase at earlier epochs when compared to KTH-TIPS-2b. The DTD dataset is more difficult (\textit{i.e.}, more images and classes) than the KTH-TIPS-2b dataset. As $d$ increased, the performance for DTD  improved as well since larger values of $d$ provide more features for the output classification layer. However, a more complex model may hinder the generalization of the model to new data as evidenced by the validation loss curves.  

\ifCLASSOPTIONcaptionsoff
  \newpage
\fi

\end{document}